\documentclass[letterpaper]{article} 
\usepackage{aaai23}  
\usepackage{times}  
\usepackage{helvet}  
\usepackage{courier}  
\usepackage[hyphens]{url}  
\usepackage{graphicx} 
\usepackage{natbib}  
\usepackage{caption} 
\usepackage{algorithm}
\usepackage{algorithmic}

\usepackage{newfloat}
\usepackage{listings}
\DeclareCaptionStyle{ruled}{labelfont=normalfont,labelsep=colon,strut=off} 
\floatstyle{ruled}
\newfloat{listing}{tb}{lst}{}
\floatname{listing}{Listing}
\usepackage{longtable}
\usepackage{array}
\usepackage{booktabs}
\usepackage{color, colortbl}
\usepackage{multirow}
\usepackage{diagbox}
\definecolor{rc1}{RGB}{235,235,235}
\definecolor{rc2}{RGB}{255,255,255}
\definecolor{fade}{gray}{0.4}
\definecolor{green}{RGB}{144,238,144}

\newcommand{\dec}[1]{\textcolor{black}{${(\downarrow#1)}$}}

\usepackage{xcolor}
\usepackage{soul}

\definecolor{fade}{gray}{0.4}

\newcommand{\specialcell}[2][c]{%
\begin{tabular}[#1]{@{}c@{}}#2\end{tabular}}
\newcommand{\specialcellleft}[2][l]{%
\begin{tabular}[#1]{@{}l@{}}#2\end{tabular}}

\usepackage{fontawesome}
\newcommand{\video}{\setlength{\fboxsep}{2pt}\colorbox{blue!30}{\faVideoCamera}}
\newcommand{\audio}{\colorbox{green!30}{\faVolumeUp}}

\usepackage{pifont}
\newcommand{\cmark}{\ding{51}}%
\newcommand{\xmark}{\ding{55}}%
\usepackage{graphicx}
\usepackage{amsmath}
\usepackage{booktabs}
\usepackage{enumitem}
\usepackage{longtable}
\usepackage{array}
\usepackage{booktabs}
\usepackage{color, colortbl}
\usepackage{multirow}
\usepackage{diagbox}
\usepackage{algorithm}
\usepackage{listings}
\definecolor{rc1}{RGB}{235,235,235}
\definecolor{rc2}{RGB}{255,255,255}
\definecolor{codeblue}{rgb}{0.25,0.5,0.5}
\definecolor{codekw}{rgb}{0.85, 0.18, 0.50}

\newcommand{\dist}{\mathcal{D}}
\newcommand{\loss}{\mathcal{L}}
\newcommand{\dash}{\texttt{-}}

\usepackage{comment}
\usepackage{slashbox}

\def\ra{{\textnormal{a}}}
\def\rv{{\textnormal{v}}}
\def\rt{{\textnormal{t}}}
\def\rT{{\textnormal{T}}}
\def\sg{{\texttt{S}}}

\usepackage{url}

\title{Self-Supervised Audio-Visual Representation Learning with Relaxed Cross-Modal Synchronicity}
\author {
    Pritam Sarkar\textsuperscript{\rm 1, \rm 2} \quad
    Ali Etemad\textsuperscript{\rm 1}
}
\affiliations {
    \textsuperscript{\rm 1} Queen's University, Canada \quad
    \textsuperscript{\rm 2} Vector Institute\\
    \texttt{\{pritam.sarkar, ali.etemad\}@queensu.ca}
    \\ {\textcolor{magenta}{\bf\small\texttt{https://pritamqu.github.io/CrissCross}}}
}

\begin{document}

\maketitle

\begin{abstract}
    We present \textbf{CrissCross}, a self-supervised framework for learning audio-visual representations. A novel notion is introduced in our framework whereby in addition to learning the intra-modal and standard `synchronous' cross-modal relations, CrissCross also learns `asynchronous' cross-modal relationships. We perform in-depth studies showing that by relaxing the temporal synchronicity between the audio and visual modalities, the network learns strong generalized representations useful for a variety of downstream tasks. To pretrain our proposed solution, we use $3$ different datasets with varying sizes, Kinetics-Sound, Kinetics400, and AudioSet. The learned representations are evaluated on a number of downstream tasks namely action recognition, sound classification, and action retrieval. Our experiments show that CrissCross either outperforms or achieves performances on par with the current state-of-the-art self-supervised methods on action recognition and action retrieval with UCF101 and HMDB51, as well as sound classification with ESC50 and DCASE. Moreover, CrissCross outperforms fully-supervised pretraining while pretrained on Kinetics-Sound. 
    The codes, pretrained models, and supplementary material are available on the project website.
\end{abstract}



\section{Introduction} \label{sec:intro}

In recent years, self-supervised learning has shown great promise in learning strong representations without human-annotated labels \cite{simclr,simsiam,deepcluster}, and emerged as a strong competitor for fully-supervised pretraining. There are a number of benefits to such methods. Firstly, they reduce the time and resources required for expensive human annotations and allow researchers to directly use large uncurated datasets for learning meaningful representations. Moreover, the models trained in a self-supervised fashion learn more abstract representations, which are useful for a variety of downstream tasks without needing to train the models from scratch. 
Given the abundance of videos, their spatio-temporal information-rich nature, and the fact that in most cases they contain both audio and visual streams, self-supervised approaches are strong alternatives to fully-supervised methods for video representation learning. Moreover, the high dimensionality and multi-modal nature of videos make them difficult to annotate, further motivating the use of self-supervision.

The common and standard practice in self-supervised audio-visual representations learning is to learn intra-modal and cross-modal relationships between the audio and visual streams by maintaining tight temporal synchronicity between the two modalities \cite{mmv,avts,xdc,selavi}. Yet, the impact of learning temporally asynchronous cross-modal relationships in the context of self-supervised learning has not been explored. This notion deserves deeper exploration as learning such temporally asynchronous cross-modal relationships may in fact result in increased invariance and distinctiveness in the learned representations.

In this study, in an attempt to explore the notion above, we present \textbf{CrissCross}, a self-supervised framework to learn robust generalized audio-visual representations from videos. CrissCross is built upon SimSiam \cite{simsiam} to jointly learn self-supervised audio-visual representations through a mixture of intra- and cross- modal optimizations. In addition to learning intra-modal and standard synchronous cross-modal relations, CrissCross introduces the novel idea of learning cross-modal representations through \textit{relaxing} time-synchronicity between corresponding audio and visual segments. We refer to this as `asynchronous cross-modal' optimization, a concept that has not been explored in prior works. We use $3$ datasets of different sizes: Kinetics-Sound \cite{l3-kineticssound}, Kinetics400 \cite{kinetics400}, and AudioSet \cite{audioset}, to pretrain CrissCross. We evaluate CrissCross on different downstream tasks, namely action recognition, sound classification, and action retrieval. We use $2$ popular benchmarks UCF101 \cite{ucf101} and HMDB51 \cite{hmdb} to perform action recognition and retrieval, while ESC50 \cite{esc} and DCASE \cite{dcase} are used for sound classification.

\noindent The key contributions of this work are as follows:
\begin{itemize}[noitemsep,nolistsep]
\item[$\bullet$]
We present a novel framework for multi-modal self-supervised learning by relaxing the audio-visual temporal synchronicity to learn effective generalized representations. 
Our method is simple, data efficient
and less resource intensive,
yet learns robust multi-modal representations for a variety of downstream tasks.
\item[$\bullet$]
We perform an in-depth study to explore the performance of the proposed framework and
its major concepts. Moreover, we perform thorough analyses, both quantitatively and qualitatively, in different setups, showing the benefit of learning asynchronous cross-modal relations.
\item[$\bullet$]
Comparing the performance of our method to prior works, CrissCross achieves state-of-the-arts on UCF101, HMDB, ESC50, and DCASE when pretrained on Kinetics400. Moreover, when trained with AudioSet, CrissCross achieves better or competitive performances versus the current state-of-the-arts.

\item[$\bullet$] Lastly, when pretrained on the small-scale Kinetics-Sound \cite{l3-kineticssound}, CrissCross outperforms fully-supervised pretraining \cite{cmacc} by $1.4\%$ and $7.4\%$, as well as prior self-supervised state-of-the-art \cite{cmacc} by $11.1\%$ and $19.9\%$ on UCF101 and HMDB51 respectively. To the best of our knowledge, very few prior works have attempted to pretrain on such small datasets, and in fact, this is the first time where self-supervised pretraining outperforms full supervision on action recognition in this setup.

\end{itemize}

We hope our proposed self-supervised method can motivate researchers to further explore the notion of \textit{asynchronous} multi-modal representation learning. 


\section{Related Work} \label{sec:related_work}
\subsection{Self-supervised Learning}
Self-supervised learning aims to learn generalized representations of data without any human annotated labels through properly designed pseudo tasks (also known as pretext tasks). Self-supervised learning has recently drawn significant attention in different areas such as image \cite{simclr,simsiam,pirl,swav,byol,deepcluster}, video \cite{avid,ravid,xdc,selavi,gdt,mmv,cmac}, and wearable data \cite{sarkar-ssl-icassp,sarkar-ssl-tafc,sarkar-ssl2} analysis among others. 

In self-supervised learning, the main focus of interest lies in designing novel pseudo-tasks to learn useful representations. We briefly mention some of the popular categories in the context of self-supervised video representation learning, namely, $i$) context-based, $ii$) generation-based, $iii$) clustering-based, and $iv$) contrastive learning-based. Various pretext tasks have been proposed in the literature exploring the spatio-temporal context of video frames, for example, temporal order prediction \cite{opn}, puzzle solving \cite{stc,shuffle-learn,ahsan2019video}, rotation prediction \cite{rotnet3d}, and others. Generation-based video feature learning methods refer to the process of learning feature representations through video generation \cite{vondrick2016generating,tulyakov2017mocogan,saito2017temporal}, video colorization \cite{tran2016deep}, and frame or clip prediction \cite{mathieu2015deep,reda2018sdc,babaeizadeh2017stochastic,liang2017dual,finn2016unsupervised}, among a few others. Clustering-based approaches \cite{xdc,selavi} rely on self-labeling where data is fed to the network and the extracted feature embeddings are clustered using a classical clustering algorithm such as k-means, followed by using the cluster assignments as the pseudo-labels for training the neural network. The key concept of contrastive learning \cite{simsiam,pirl,byol,swav,avid,gdt} is that in the embedding space, `positive' samples should be similar to each other, and `negative' samples should have discriminative properties. Using this concept, several prior works \cite{avid,ravid,gdt,cmacc} have attempted to learn representations by minimizing the distance between positive pairs and maximizing the distance between negative pairs.

\subsection{Audio-Visual Representation Learning}

Typically in multi-modal self-supervised learning, multiple networks are jointly trained on the pseudo tasks towards maximizing the mutual information between multiple data streams \cite{xdc,avid,avts,cliporder,wang2021self,khare2021self,siriwardhana2020multimodal}. Following, we briefly discuss some of the prior works \cite{avts,xdc,avid,cmacc} on audio-visual representation learning. 
A multi-modal self-supervised task is introduced in AVTS~\cite{avts}, leveraging the natural synergy between audio-visual data. The network is trained to distinguish whether the given audio and visual sequences are `in sync' or `out of sync'. 
In XDC~\cite{xdc}, the authors introduce a framework to learn cross-modal representations through a self-labeling process. Specifically, cross-modal pseudo-labeling is performed where the pseudo-labels computed from audio embeddings are used to train the visual backbone, while the pseudo-labels computed using visual embeddings are used to train the audio network. A self-supervised learning framework based on contrastive learning is proposed in AVID~\cite{avid} to learn audio-visual representations from videos. AVID performs instance discrimination as the pretext task by maximizing the cross-modal agreement of the audio-visual segments in addition to visual similarity. Though earlier works focus on learning cross-modal relations while maintaining a tight synchronicity between the audio and visual data, our proposed framework also considers asynchronous cross-modal relationships in addition to the standard synchronous relations.


\section{Method} \label{sec:method}

\subsection{Approach}
Let be given $v$, a sequence of visual frames, and $a$, the corresponding audio waveform. We can obtain $n$ augmented views of $v$ as $\{v_i\}_{i=0}^{n}$, and equal number of augmented views of $a$ as $\{a_i\}_{i=0}^{n}$. A common way to learn individual representations from $v$ and $a$ is to minimize the embedding distances ($\dist$) between the augmented views of the each modality as $\loss_{vv}\!=\!\sum_{i,j=0, i\neq j}^{n} \dist(v_i, v_j)$ and $\loss_{aa}\!=\!\sum_{i,j=0, i\neq j}^{n} \dist(a_i, a_j)$ respectively in a self-supervised setting \cite{swav,vicreg,simsiam,byol,byol_audio}. Further, to learn multi-modal representations from $\{v, a\}$, a standard technique is to simply optimize a joint intra-modal loss $\loss_{intra} = \loss_{vv} + \loss_{aa}$. Prior works \cite{xdc,avid,ravid} have demonstrated that in addition to $\loss_{intra}$, a cross-modal optimization can be performed directly across visual and audio segments to further learn strong joint representations as
$\loss_{av}\!=\!\sum_{i=0}^{n} \dist(a_i, v_i)$.

\begin{figure}[!t]
    \centering

    \includegraphics[width=0.65\linewidth]{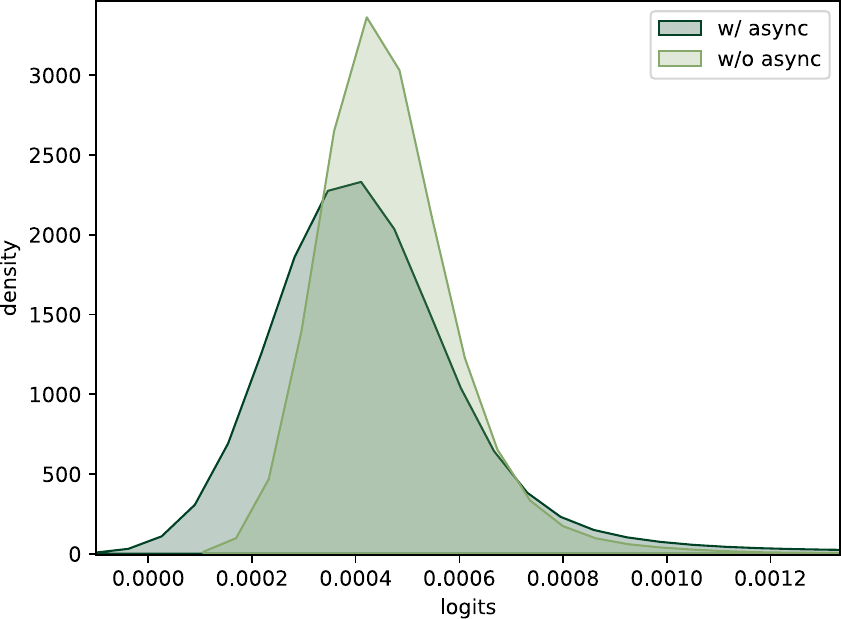}
    \caption{Distribution of the learned representations with and without the asynchronous cross-modal optimization.}

    \label{fig:dist_plot}
    
    \vspace{10pt}
    \includegraphics[width=0.9\linewidth]{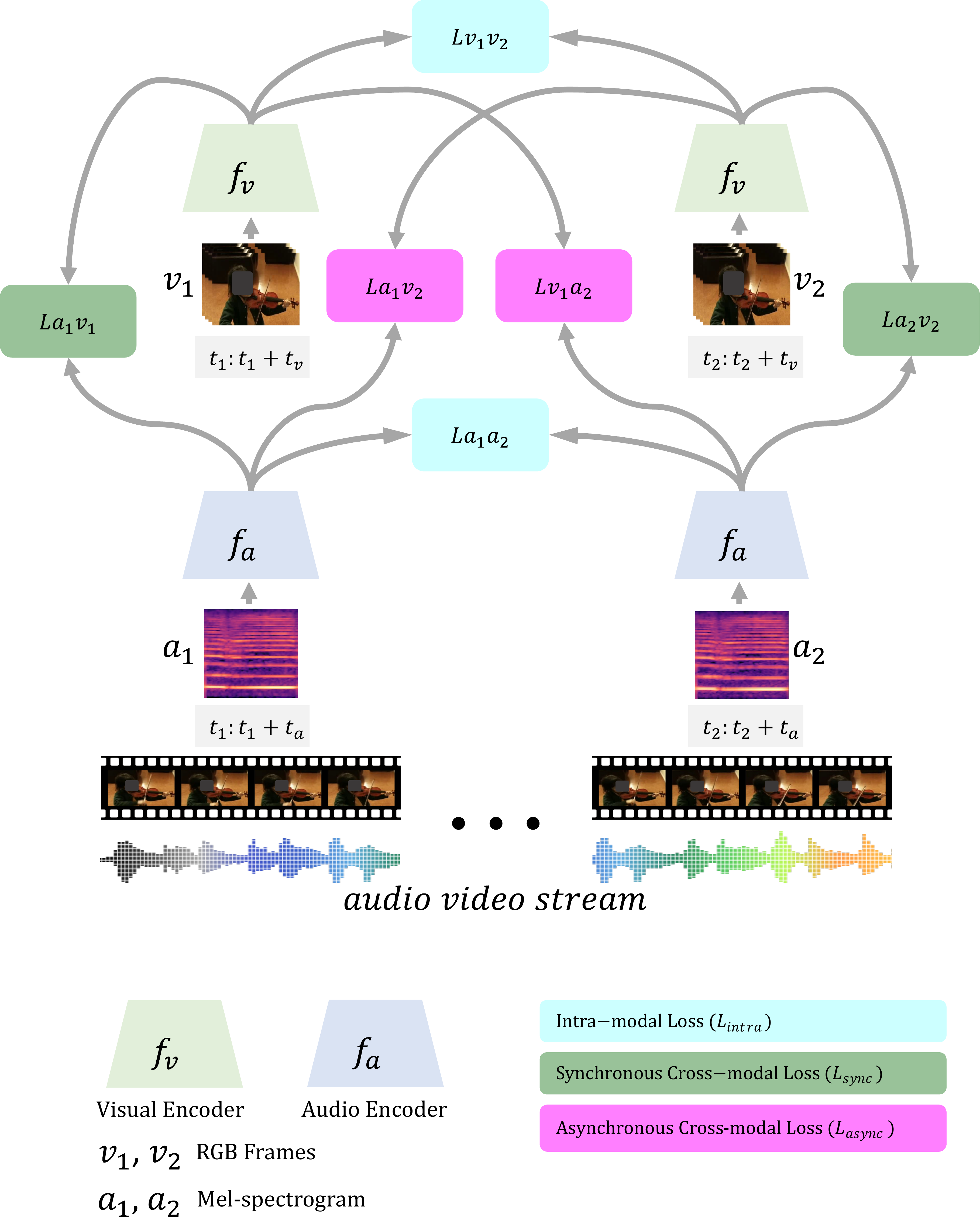}
    \caption{{Our proposed framework.} CrissCross learns strong audio-visual representations by exploiting  intra-modal, as well as, sync. and async. cross-modal relations. }

    \label{fig:framework}
    
\end{figure}

All of these learning procedures maintain a tight synchronicity between the two modalities, given that both $a_i$ and $v_i$ are segmented from the same timestamps. We conjecture, however, that \textit{relaxing the synchronicity between modalities by a reasonable margin will enable more generalized representations to be learned} across time, to achieve better and more robust performance. Accordingly, we introduce asynchronous cross-modal loss $\loss_{async}$, which exploits the relationship between audio and visual segments sampled at different timestamps. We define the final objective as $\loss_{CrissCross}$ which exploits the combination of $\loss_{intra}$, synchronous $\loss_{av}$ (which we refer to as $\loss_{sync}$), and $\loss_{async}$ in an attempt to learn more generalized representations. While we present the detailed experiments and analysis of our proposed approach in the subsequent sections of the paper, here we perform a quick visualization to demonstrate the benefits of this concept. Figure \ref{fig:dist_plot} depicts the distributions of representations learned with and without $\loss_{async}$, demonstrating that indeed relaxing the tight synchronicity helps in widening the distribution of the learned representations which could result in improved performance in a wide variety of downstream tasks.

\subsection{Training Objective}

To accomplish the notion above, let's define two neural networks, a visual encoder $f_v$ and an audio encoder $f_a$. Here, $f_v$ and $f_a$ are composed of convolutional backbones and MLP projection heads. Moreover, we adopt a Siamese \cite{bromley1993signature} representation learning setup, where the networks share weights on two or more inputs. Next, We obtain two augmented views of $v\!=\!\{\rv_\rt\}_{\rt=0}^{\rT}$, denoted by $v_1$ and $v_2$, defined as $\{\rv_\rt\}_{\rt=\rt_1}^{\rt_1+\rt_v}$ and $\{\rv_\rt\}_{\rt=\rt_2}^{\rt_2+\rt_v}$ respectively. Here, $v_1$ and $v_2$ have a duration of $\rt_v$, and are sampled at times $\rt_1$ and $\rt_2$ respectively. Note that $v_1$ and $v_2$ are augmented differently. Similarly, two augmented views of $a\!=\!\{\ra_\rt\}_{\rt=0}^{\rT}$ can be obtained as $a_1$ and $a_2$ as $\{\ra_\rt\}_{\rt=\rt_1}^{\rt_1+\rt_a}$ and $\{\ra_\rt\}_{\rt=\rt_2}^{\rt_2+\rt_a}$, respectively. Next, to learn intra-modal representations, the distance between $f_v(v_1)$ and $f_v(v_2)$, as well as, $f_a(a_1)$ and $f_a(a_2)$ can be minimized to train $f_v$ and $f_a$ respectively. However, such a naive approach would lead to mode collapse as pointed out in \cite{byol,byol_audio,simsiam,swav}. To tackle this, we follow the technique proposed in \cite{simsiam}. In particular, we minimize the cosine embedding distance $\dist$ of two output vectors $p$ and $\sg(z)$, where $p$ is the output vector obtained from the predictor head and $z$ represents the output vector obtained from the feature encoder followed by the \texttt{stop-gradient} operation. Here, the predictor head consists of an MLP head, which is used as an identity mapping, while the \texttt{stop-gradient} operation prevents the model from collapsing to a degenerated solution \cite{simsiam}. Here, $\dist$ is defined as:
\begin{equation}
\begin{split}
    \dist(p, z) = - \frac{p}{||p||_2} \cdot \frac{z}{||z||_2} ~.
\end{split}
\end{equation}
We use $h_v$ and $h_a$ as the predictor heads corresponding to visual and audio representations. Next, we obtain $p_{v_1}$ and $z_{v_2}$ as $h_v(f_v(v_1))$ and $\sg(f_v(v_2))$. Similarly, $p_{a_1}$ and $z_{a_2}$ are obtained as $h_a(f_a(a_1))$ and $\sg(f_a(a_2))$. To calculate the symmetrized loss, we further obtain $p_{v_2}$ and $z_{v_1}$, as well as, $p_{a_2}$ and $z_{a_1}$. 
Therefore, to learn the intra-modal relations, we optimize the intra-modal loss $\loss_{intra}$ defined as:
\begin{equation}
\begin{split}
    \loss_{intra} = (\frac{1}{2}\dist(p_{v_1}, \sg(z_{v_2})) + \frac{1}{2}\dist(p_{v_2}, \sg(z_{v_1})) \\
      + \frac{1}{2}\dist(p_{a_1}, \sg(z_{a_2})) + \frac{1}{2}\dist(p_{a_2}, \sg(z_{a_1})))/2 ~.
\end{split}
\end{equation}
Next, to learn synchronous cross-modal relations, we optimize the synchronous cross-modal loss $\loss_{sync}$, defined as:
\begin{equation}
\begin{split}
    \loss_{sync} = (\frac{1}{2}\dist(p_{v_1}, \sg(z_{a_1})) + \frac{1}{2}\dist(p_{a_1}, \sg(z_{v_1})) \\
      + \frac{1}{2}\dist(p_{v_2}, \sg(z_{a_2})) + \frac{1}{2}\dist(p_{a_2}, \sg(z_{v_2})))/2 ~. 
\end{split}
\end{equation}
Additionally, based on our earlier intuition, to relax the temporal synchronicity, we minimize the distance between the audio and visual segments originated from different timestamps. We define asynchronous cross-modal loss $\loss_{async}$ as:
\begin{equation}
\begin{split}
    \loss_{async} = (\frac{1}{2}\dist(p_{v_1}, \sg(z_{a_2})) + \frac{1}{2}\dist(p_{a_2}, \sg(z_{v_1})) \\
      + \frac{1}{2}\dist(p_{v_2}, \sg(z_{a_1})) + \frac{1}{2}\dist(p_{a_1}, \sg(z_{v_2})))/2 ~.
\end{split}
\end{equation}
Finally, to exploit intra-modal, as well as, synchronous and asynchronous cross-modal relations we define the final objective function as:
\begin{equation}
\begin{split}
    \loss_{CrissCross} = \frac{1}{3}(\loss_{intra} + \loss_{sync} + \loss_{async}) ~.
\end{split}
\end{equation}
We present the proposed CrissCross framework in Figure \ref{fig:framework}. Please note, for the sake of simplicity, we omit showing the stop-grad and predictor head connections in Figure \ref{fig:framework}. We present the pseudocode in Appendix \ref{supsec:algo}. 


\section{Experiments and Results} \label{sec:experiment}

The details of the experiment setup and the findings of our thorough ablation studies investigating the major concepts of our proposed framework are presented here. Additionally, we extensively investigate a wide range of audio-visual augmentation techniques capable of learning strong audio-visual representations within our framework, the details are as follows. %

\subsection{Experiment Setup} 

Following the standard practice among the prior works \cite{avid,xdc,selavi,gdt,cmacc}, we use Kinetics-Sound, Kinetics400, and AudioSet for pretraining. Additionally, Kinetics400, UCF101, HMDB51, ESC50 and DCASE are used for downstream evaluation. 
We use R(2+1)D \cite{r2plus1d} and ResNet \cite{resnet} as the visual and audio backbones. 
To pretrain the network in a self-supervised fashion with audio-visual inputs, we downsample the visual streams to $16$ frames per second and feed $8$ frames of resolution $112^2$ to the visual encoder. 
Next, we downsample the audio signals to $16$kHz, and segment them into $2$-second segments. We transform the segmented raw audio waveforms to mel-spectrograms using $80$ mel filters, we set the hop size as $10$ milliseconds and FFT window length as $1024$. Finally, we feed spectrograms of shape  $80\times200$ to the audio encoder. We use Adam \cite{adam} optimizer with a cosine learning rate scheduler \cite{cosine_lrs} to pretrain the encoders and use a fixed learning rate to train the predictors. 
Please note that during the design exploration, we use Kinetics-Sound for pretraining, while the downstream evaluations are performed on UCF101 and ESC50 unless stated otherwise. We perform linear evaluations using $8$ frames of visual input and $2$ seconds of audio input. Next, a linear SVM classifier is trained using the extracted features, and report the top-1 accuracy for sample-level predictions. We provide the additional details of the experiment setup, datasets, architectures, and evaluation protocols in the Appendix.


\begin{table}[t]
    \fontsize{9pt}{10pt}\selectfont
    \centering
    \begin{tabular}{lll}
    \toprule
    \multirow{1}{*}{\textbf{Method}} &
    \textbf{UCF101} & \textbf{ESC50} \\ \midrule\midrule
    $\loss_{v_1v_2}$  & $69.1$ & \texttt{-} \\
    $\loss_{a_1a_2}$  & \texttt{-} & $62.0$ \\ 
    $\loss_{intra}$  & $69.7$ & $71.8$ \\ 
    $\loss_{sync}$  & $70.1$ & $75.8$ \\ 
    $\loss_{async}$  & $69.1$ & $74.8$ \\ 
    $\loss_{sync}+\loss_{intra}$  & $73.8$ & $78.0 $ \\ 
    $\loss_{sync}+\loss_{async}$  & $69.1 $& $74.8$ \\ 
    $\loss_{async}+\loss_{intra}$  & $72.4$ & $75.3 $\\ 
    $\loss_{v_1v_2}+\loss_{sync}+\loss_{async}$  & $71.3$ & $78.5$ \\ 
    $\loss_{a_1a_2}+\loss_{sync}+\loss_{async}$  & $70.8$ & $75.3$ \\ 

    $\mathbf{\loss_{CrissCross}}$  & $\mathbf{74.8}$ &  $\mathbf{79.0}$ \\
    \bottomrule
    \end{tabular}%
    \caption{
    We present the top-$1$ accuracy of CrissCross and its ablation variants, pretrained on Kinetics-Sound.
        }
    \label{tab:abl_variant_ks}
\end{table}

\begin{table}[t]
    \centering
    \fontsize{9pt}{10pt}\selectfont
    \begin{tabular}{llcc}
    \toprule
     \textbf{Pretrain} & \textbf{Downstream} & \specialcell{\textbf{w/o} {$\mathbf{\loss_{async}}$}} & \specialcell{\textbf{w/} {$\mathbf{\loss_{async}}$}} \\ \midrule\midrule
     KS & UCF101 & $73.8$\dec{1.0} & $\mathbf{74.8}$ \\
     KS & ESC50 & $78.0$\dec{1.0} & $\mathbf{79.0}$ \\ \midrule
     K400 & UCF101 & $75.8$\dec{4.1} & $\mathbf{79.9}$ \\
     K400 & ESC50 & $78.5$\dec{3.5} & $\mathbf{82.0}$ \\ \midrule
     K400 & KS (a) & $43.2$\dec{3.9} & $\mathbf{47.1}$ \\
     K400 & KS (v) & $53.3$\dec{2.4} & $\mathbf{55.7}$  \\
     K400 & KS (a+v) & $65.0$\dec{1.7} & $\mathbf{66.7}$ \\ 
    \bottomrule
    \end{tabular}%
    \caption{Impact of $\loss_{async}$ optimization in different pretraining and evaluation setups.
    Here, K400: Kinetics400, KS: Kinetics-Sound.
    }
    \label{tab:fusion}
\end{table}

\subsection{Ablation Study} \label{sec:ablation}
We present the ablation results in Tables \ref{tab:abl_variant_ks} and \ref{tab:fusion} to show the improvements made by optimizing asynchronous cross-modal loss in addition to intra-modal and synchronous cross-modal losses. 
First, using Kinetics-Sound, we train the framework in uni-modal setups, denoted as $\loss_{v_1v_2}$ and $\loss_{a_1a_2}$. We report the top-1 accuracy of UCF101 and ESC50 as $69.1\%$ and $62.0\%$ respectively. Next, we train the network in a multi-modal setup, where we find that $\loss_{sync}$ outperforms the other multi-modal variants including $\loss_{intra}$ and $\loss_{async}$, as well as, uni-modal baselines $\loss_{v_1v_2}$ and $\loss_{a_1a_2}$. Further study shows that combining all the multi-modal losses improves the model performance. $\loss_{CrissCross}$ outperforms $\loss_{sync}$ by $4.7\%$ and $3.2\%$ on action recognition and sound classification, respectively.

Further, to study the effect of $\loss_{async}$ in particular, we perform ablation studies using small-scale Kinetics-Sound and large-scale Kinetics400. We present the results in Table \ref{tab:fusion}, where we observe that $\loss_{async}$ improves the performance on both the pretraining datasets. In particular, while pretrained on Kinetics400, optimizing $\loss_{async}$ in addition to $\loss_{sync}$ and $\loss_{intra}$ improves the performances by ${4.1\%}$ and ${3.5\%}$ on action recognition and sound classification respectively, showing the significance of asynchronous cross-modal optimization in a multi-modal setup. While pretrained on Kinetics-Sound, adding $\loss_{async}$ improves the performances
by $1\%$ on both the UCF101 and ESC50. We interestingly find that learning asynchronous cross-modal loss significantly improves the model performance when pretrained on large-scale Kinetics400. Our intuition is that as Kinetics-Sound consists of a few hand-picked classes which are prominently manifested in both audio and visual modalities, the performance gain due to $\loss_{async}$ is less prominent. However, Kinetics400 is considerably larger in scale and comprises highly diverse action classes which are not always very prominent both audibly and visually. It therefore benefits more from the generalized representations learned by asynchronous cross-modal optimization. Moreover, to demonstrate the benefit of optimizing $\loss_{async}$ throughout the pretraining process, we present the top-1 accuracy vs. pretraining epoch in Figure \ref{fig:pretr_epoch_vs_acc_abl_temp_sampling}. It shows that $\loss_{async}$ significantly improves the model performance throughout the pretraining. 

\subsubsection{Multi-modal fusion.} Next, we investigate if learning asynchronous cross-modal relations helps in multi-modal fusion. To test this, we use Kinetics-Sound as the downstream dataset and Kinetics400 as the pretraining dataset. We choose Kinetics-Sound for downstream evaluation as it consists of action classes that are represented prominently in both audio and visual domains. The results are presented in Table \ref{tab:fusion}, where it is shown that learning asynchronous cross-modal relations improves multi-modal fusion by $1.7\%$. 
Additionally, we show the linear evaluation results obtained from the uni-modal feature representations for reference. It shows that optimizing $\loss_{async}$ improves the action classification accuracy by $2.4\%$ and $3.9\%$ using visual and audio representations, respectively.

\subsubsection{Qualitative analysis.} Lastly, to perform a qualitative analysis on the impact of $\loss_{async}$ we visualize the saliency maps obtained from the models when pretrained with and without the presence of the asynchronous loss. In this experiment, we directly use the models pretrained on Kinetics400 and use Grad-CAM
\cite{omeiza2019smooth} to visualize randomly selected samples from Kinetics400. A few examples are presented in Figure \ref{fig:async_visual}, where we observe that learning asynchronous relations helps the model focus better on the salient information. 
Specifically, we notice that optimizing $\loss_{async}$ helps in correctly locating the sound sources on the visual streams, as shown by the examples of `dribbling basketball', `laughing', `tapping guitar', etc. 

\begin{table}[t]
    \centering
    \resizebox{0.45\textwidth}{!}{
    \begin{tabular}{cc}
    \textbf{w/o asynchronous loss} & \textbf{w/ asynchronous loss} \\
         \includegraphics[width=\linewidth]{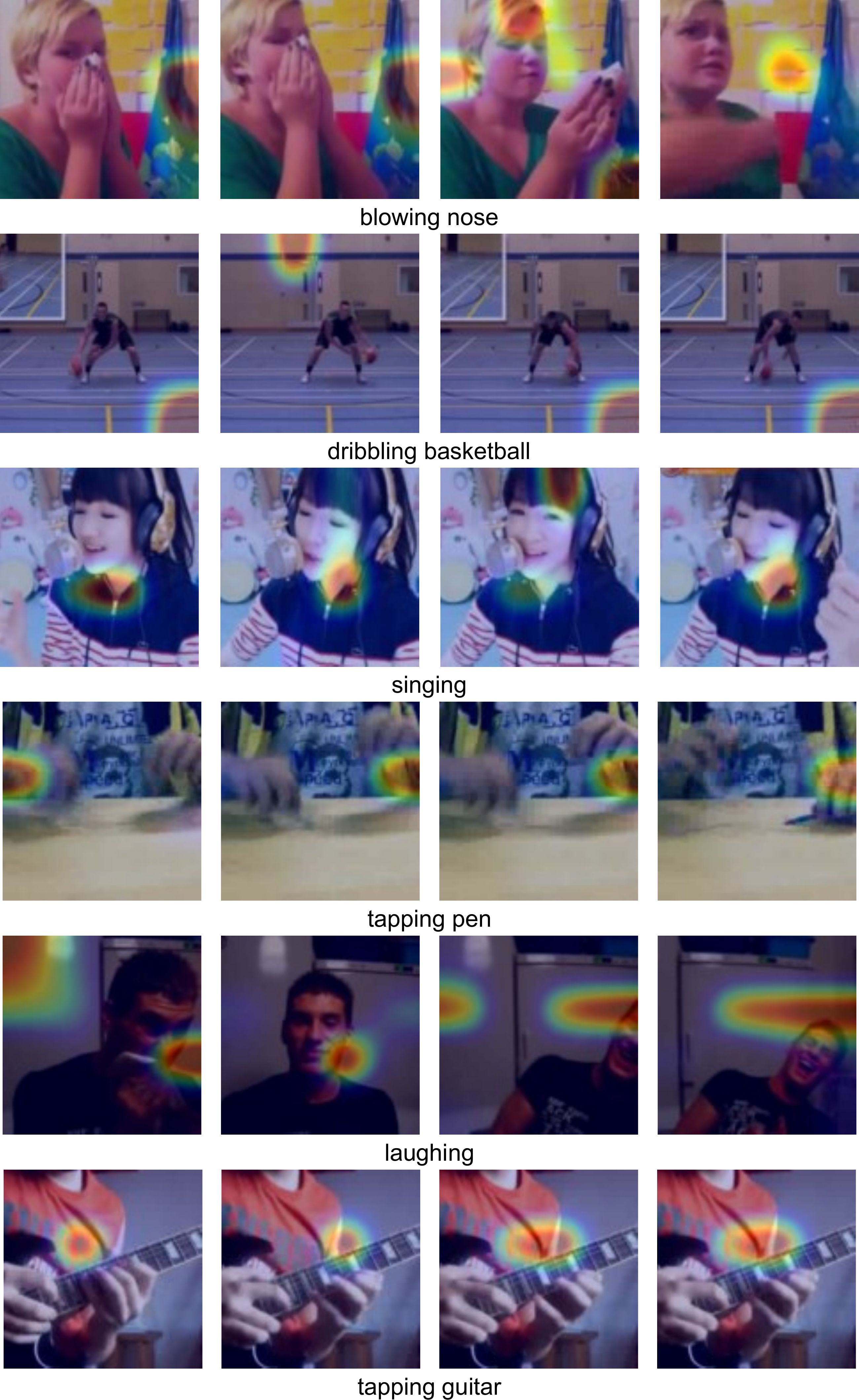}
         &  
         \includegraphics[width=\linewidth]{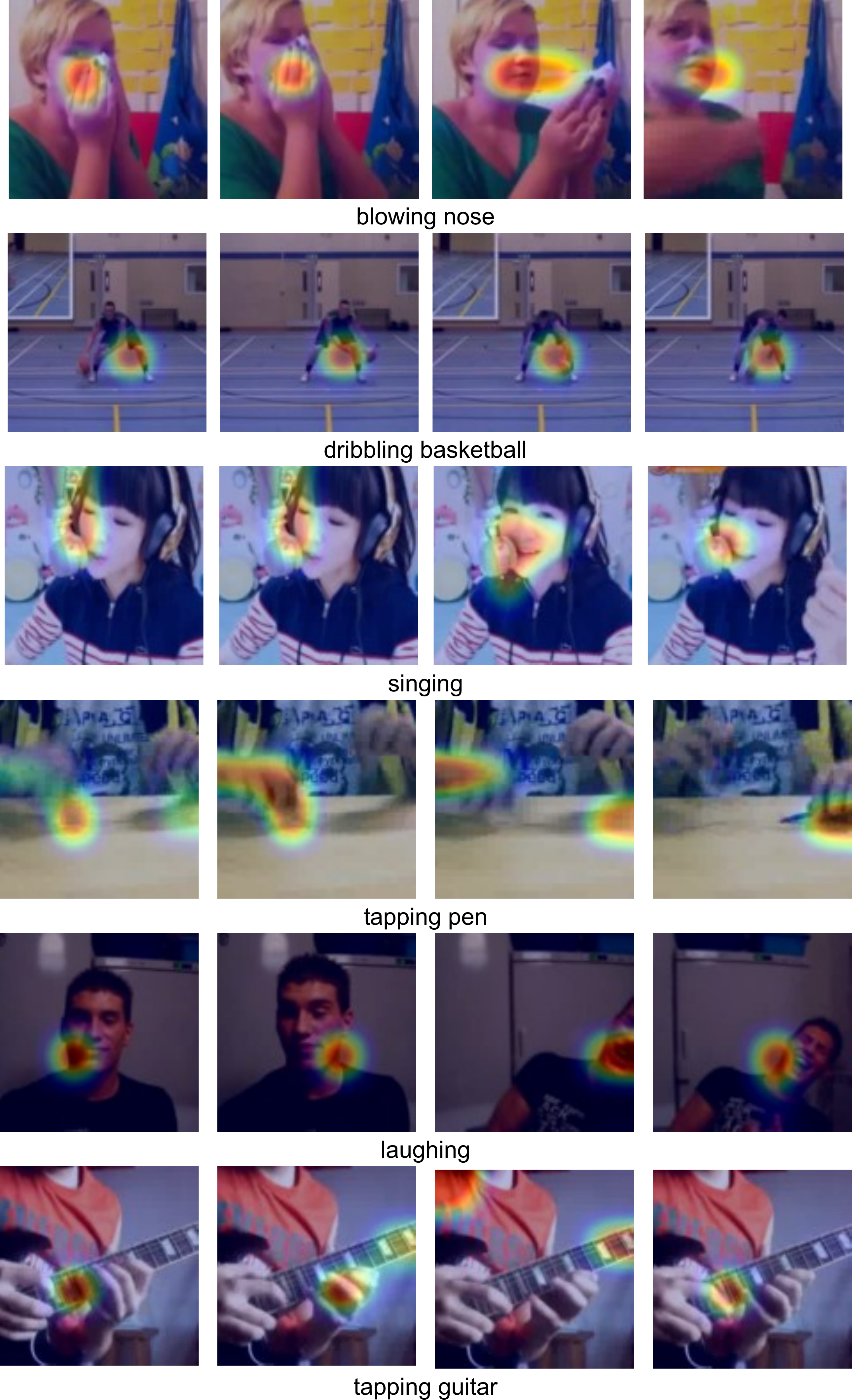}
         \\
    \end{tabular}
    }
    \captionof{figure}{Visualization of saliency maps while pretrained without (left) and with (right) asynchronous loss. 
    }
    \label{fig:async_visual}
    
    \vspace{10pt}
    
    \begin{tabular}{cc}
     \includegraphics[width=0.45\columnwidth]{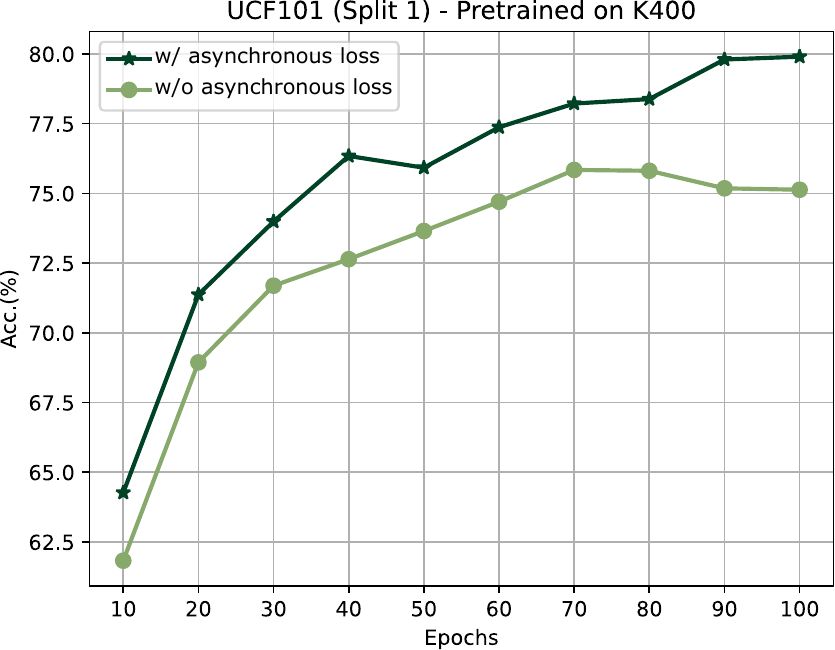}
    &
    \includegraphics[width=0.45\columnwidth]{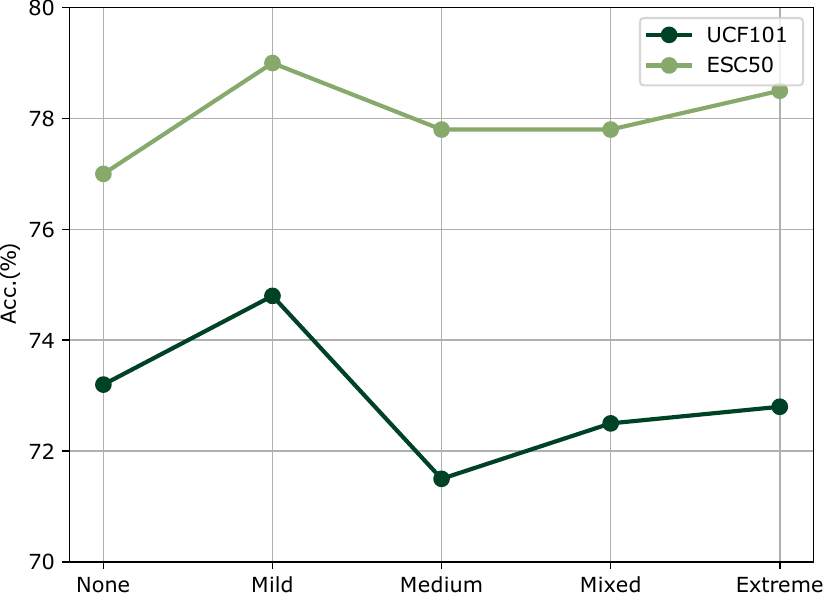}
    \end{tabular}
    \captionof{figure}{Left: Linear eval. top-1 acc. vs. pretraining epochs.
    Right: Exploring different temporal relaxation techniques.}
    \label{fig:pretr_epoch_vs_acc_abl_temp_sampling}
\end{table}

\subsection{Exploring Relaxed Time-synchronicity}
Audio and visual modalities from the same source clip generally maintain a very strong correlation, which makes them suitable for multi-modal representation learning as one modality can be used as a supervisory signal for the other in a self-supervised setup. However, our intuition behind CrissCross is that these cross-modal temporal correlations do not necessarily need to follow a strict frame-wise coupling. Instead, we hypothesize that relaxing cross-modal temporal synchronicity to some extent can help in learning more generalized representations.

To facilitate this idea within CrissCross, we exploit $5$ different temporal sampling methods to explore varying amounts of {temporal synchronicity} when learning cross-modal relationships. 
(\textit{i}) \textit{None:} where both the audio and visual segments are sampled from the exact same time window. (\textit{ii}) \textit{Mild:} where the two views of the audio-visual segments share $50\%$ overlap amongst them. (\textit{iii}) \textit{Medium:} where adjacent frame sequences and audio segments are sampled. (\textit{iv}) \textit{Extreme:} in which we sample one view from the first half of the source clip, while the other view is sampled from the second half of the source clip. (\textit{v}) \textit{Mixed:} where the two audio-visual segments are sampled in a temporally random manner.
The results presented in Figure \ref{fig:pretr_epoch_vs_acc_abl_temp_sampling} show that the \textit{mild} relaxation works best for both action recognition and sound classification. Interestingly, we find that \textit{medium} relaxation shows worse performance in comparison to others, whereas, \textit{extreme} relaxation works somewhat well in our setup.

\begin{table}[t]
    \centering
    \resizebox{0.99\linewidth}{!}{%
        \begin{tabular}{lcccc}
        \toprule
         & \textbf{lr}$\mathbf{_{\texttt{p}}\!=\!}$ \textbf{lr}$\mathbf{_{\texttt{b}}}$ & \textbf{comm. pred.} & \textbf{2 layers proj.} & \textbf{default}
        \\  \midrule\midrule
        UCF101 & $59.0$ & $73.6$ & $72.4$ & $\mathbf{74.8}$ \\
        ESC50 & $62.3$ & $75.3$ & $75.0$ &  $\mathbf{79.0}$ \\
        \bottomrule
        \end{tabular}%
        }
    \caption{A comparative study of different predictor and projector setups.
         Here, lr$_{\texttt{b}}$: base LR and lr$_{\texttt{p}}$: pred LR
        }
    \label{tab:abl_pred_proj}
\end{table}

\begin{table}[t]
\fontsize{9pt}{10pt}\selectfont
    \centering
    \begin{tabular}{c}
    \includegraphics[width=0.9\linewidth]{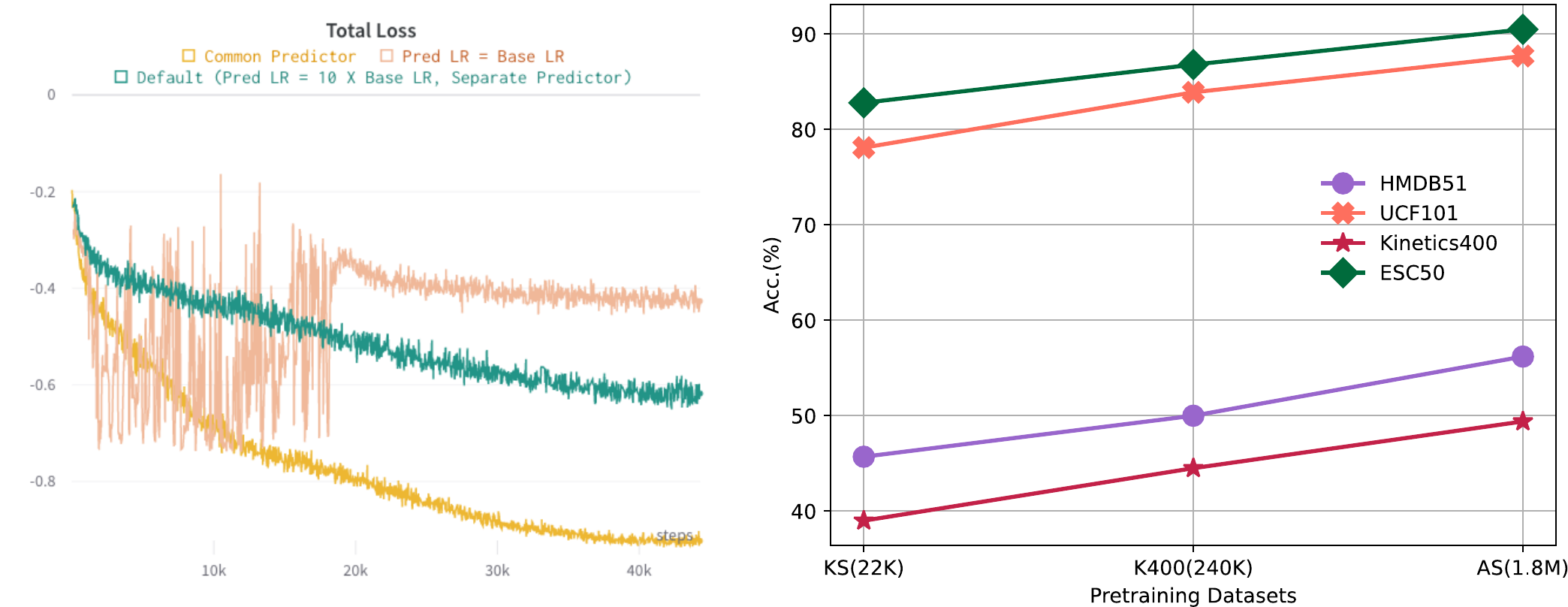}
    \end{tabular}
    
    \captionof{figure}{Left: Pretraining loss curves during predictor head design exploration.
    Right: Linear evaluation top-1 accuracy vs. pretraining dataset (size).}
    \label{fig:loss_lr_pred}
\end{table}

\begin{table}[t]
\fontsize{9pt}{10pt}\selectfont
    \centering    
    \begin{tabular}{lccc}
    \toprule
    & \multicolumn{3}{c}{\textbf{Pretraining Dataset}} \\ \cmidrule{2-4}
    & \specialcell{\textbf{KS (22K)}} & \specialcell{\textbf{K400 (240K)}} & \specialcell{\textbf{AS (1.8M)}} \\ \midrule\midrule
    HMDB51 &$45.7$&$50.0$&$56.2$\\
    UCF101 &$78.1$&$83.9$&$87.7$\\
    Kinetics400 &$39.0$ &$44.5$&$50.1$\\
    ESC50 &$82.8$&$86.8$&$90.5$ \\
    DCASE &$93.0$&$96.0$&$97.0$\\
    \bottomrule
    \end{tabular}
    \captionof{table}{We present the top-1 acc. of {linear} evaluation on action recognition and sound classification.
    }
    \label{tab:linear_results_new}

\end{table}

\subsection{Exploring Design Choices}
\subsubsection{Predictor.}
Our empirical study shows that the predictor head plays an important role in effectively training the audio and visual encoders to learn good representations. The predictor architecture is similar to \cite{simsiam}. For the sake of completeness, we provide the details of the predictor head in Appendix \ref{supsec:arch}. We explore (\textit{i}) different learning rates, and (\textit{ii}) using a common vs. a separate predictor in the multi-modal setup. It should be noted that none of the variants cause a collapse, even though we notice considerable differences in performance. We present the findings below. 

Following \cite{simsiam}, we use a constant learning rate for the predictors. However, unlike \cite{simsiam}, where the predictor learning rate is the same as the base learning rate of the encoder, we find that a higher predictor learning rate helps the network to learn better representations. In particular, setting the predictor learning rate to be the same as the base learning rate results in unstable training, and the loss curve shows oscillating behavior. We empirically find that setting the predictor learning rate to $10$ times the base learning rate works well. We present the results in Table \ref{tab:abl_pred_proj} and training curves in Figure \ref{fig:loss_lr_pred}.

Next, we evaluate whether the framework can be trained with a common predictor head instead of separate predictor heads (default setup). In simple terms, one predictor head would work towards identity mapping for both audio and visual feature vectors. To test this, l2-normalized feature vectors $f_v(v)$ and $f_a(a)$ are fed to the predictor, which are then used in a usual manner to optimize the cost function. The results are presented in Table \ref{tab:abl_pred_proj}. We observe that though such a setup works somewhat well, having separate predictors is beneficial for learning better representations. We present the training curves in Figure \ref{fig:loss_lr_pred}, it shows using common predictor head results in training losses saturate very quickly ultimately yielding worse performance compared to the use of separate predictor heads.

\subsubsection{Projector.}
We present a comparative study of projection heads with $2$ layers vs. $3$ layers (default setup). We notice $2.4\%$ and $4\%$ improvements in top-1 accuracies when using $3$ layers instead of $2$ on action recognition and sound classification respectively (please see Table \ref{tab:abl_pred_proj}). The architecture details are presented in Appendix \ref{supsec:arch}.

\subsection{Exploring Audio-Visual Augmentations.}
We perform an in-depth study to explore the impact of different audio and visual augmentations. 

\noindent\textbf{Visual Augmentations.}
We explore a wide range of visual augmentations. As a starting point, we adopt the basic spatial augmentations used in \cite{avid}, which consists of Multi-Scale Crop (MSC), Horizontal Flip (HF), and Color Jitter (CJ). Additionally, we explore other augmentations, namely Gray Scale (GS), Gaussian Blur (GB) \cite{simclr}, and Cutout (C) \cite{cutout}, which show great performance in image-based self-supervised learning \cite{simclr,scan}. We explore almost all the possible combinations of different visual augmentations in a uni-modal setup and present the results in Table \ref{tab:abl_aud_vid_aug}. The results show that strong augmentations improve the top-1 accuracy by $6.8\%$ in comparison to basic augmentations used in \cite{avid}.

\noindent\textbf{Temporal Consistency of Spatial Augmentations.}
While investigating different spatial augmentations, we are also interested to know if the spatial augmentations should be consistent at the frame level or whether they should be random (i.e., vary among consecutive frames within a sequence). We refer to these concepts as \textit{temporarily consistent} or \textit{temporarily random}. We perform an experiment where we apply MSC\dash HF\dash CJ\dash GS randomly at the frame level and compare the results to applying the same augmentations consistently across all the frames of a sequence. Our results show that maintaining temporal consistency in spatial augmentations across consecutive frames is beneficial, which is in line with the findings in \cite{cvrl}. Specifically, \textit{Temporally random} augmentations, results in top-1 accuracy of $53.69\%$, whereas, the same augmentations applied in a \textit{temporally consistent} manner results in $68.09\%$.

\noindent\textbf{Audio Augmentations.}
Similar to visual augmentations, we thoroughly investigate a variety of audio augmentations. Our audio augmentations include, Volume Jitter (VJ), Time and Frequency Masking (Mask) \cite{specaug}, Random Crop (RC) \cite{byol_audio}, and Time Warping (TW) \cite{specaug}. We also explore almost all the possible combinations of these augmentations and present the results in Table \ref{tab:abl_aud_vid_aug}. Our findings show that time-frequency masking and random crop improve the top-1 accuracy by $17.25\%$ compared to the base variant. We also notice that time warping doesn't improve performance and is also quite computationally expensive. Hence, going forward we do not use time warping during pretraining.

\noindent\textbf{Audio-Visual Augmentations.}
We conduct further experiments on a few combinations of augmentations in a \textit{multi-modal} setup. We pick the top-performing augmentations obtained from the uni-modal variants and apply them concurrently. The results are presented in Table \ref{tab:abl_aud_vid_aug} where we find that the results are consistent with the uni-modal setups, as the combination of MSC\dash HF\dash CJ\dash GS\dash GB\dash C and VJ\dash M\dash RC performs the best in comparison to the other combinations. Finally, We summarize the augmentation schemes used for pretraining and evaluation in Tables \ref{stab:summ_vid_aug} and \ref{stab:summ_aud_aug}.

\begin{table}[t]
    \fontsize{9pt}{10pt}\selectfont
        \centering
        
        \resizebox{0.47\textwidth}{!}{%
        \begin{tabular}{llllll}
        \toprule
        & \textbf{Visual} & \textbf{UCF101} && \textbf{Audio} & \textbf{ESC50} \\ \cmidrule{1-6}
        \parbox[t]{2mm}{\multirow{5}{*}{\rotatebox[origin=c]{90}{Uni}}}
        & MSC\dash HF\dash CJ & 62.3 && {VJ} & 44.8 \\
        & MSC\dash HF\dash CJ\dash GS & 68.1 && {VJ\dash M} & 49.5 \\
        & MSC\dash HF\dash CJ\dash GS\dash C & 68.3 && {VJ\dash M\dash TW} & 49.5 \\
        & MSC\dash HF\dash CJ\dash GS\dash GB & 68.7 && {VJ\dash M\dash RC} & $\mathbf{62.0}$ \\
        & MSC\dash HF\dash CJ\dash GS\dash GB\dash C & $\mathbf{69.1}$ &&  &  \\
        \toprule
        & \multicolumn{2}{l}{\textbf{Visual + Audio}} && \textbf{UCF101} & \textbf{ESC50} \\ \cmidrule{1-6}
        \parbox[t]{2mm}{\multirow{3}{*}{\rotatebox[origin=c]{90}{Multi}}}
        & \multicolumn{2}{l}{MSC\dash HF\dash CJ\dash GS\dash C + VJ\dash M\dash RC} && $73.9$ & $79.0$ \\ 
        & \multicolumn{2}{l}{MSC\dash HF\dash CJ\dash GS\dash GB + VJ\dash M\dash RC} && $73.5$ & $79.0$ \\ 
        & \multicolumn{2}{l}{MSC\dash HF\dash CJ\dash GS\dash GB\dash C + VJ\dash M\dash RC} && $\mathbf{74.8}$ & $\mathbf{79.0}$ \\
        \bottomrule
        \end{tabular}%
        }
        \caption{Exploring audio-visual augmentations.}
        \label{tab:abl_aud_vid_aug}
    \end{table}%

\subsection{Linear Evaluation and Scalability} \label{sec:linear_eval}
To evaluate the quality of the representations learned through pretraining, we perform linear evaluation on action recognition (HMDB51, UCF101, and Kinetics400) and sound classification (ESC50 and DCASE). As mentioned, we use $3$ different-sized datasets, i.e., Kinetics-Sound, Kinetics400, and AudioSet for pretraining. In Table \ref{tab:linear_results_new} we report the top-1 accuracies averaged over all the splits.
Moreover, to evaluate the scalability of CrissCross, we plot the linear evaluation results against the size of pretraining data as shown in Figure \ref{fig:loss_lr_pred}. We notice a steady improvement in performance as the dataset size increases, which shows CrissCross can likely be scaled on even larger datasets like IG65M \cite{ig65m}. 
Please note that in order to evaluate scalability we choose linear evaluation accuracy instead of full-finetuning as it gives more accurate measurements of learned representations obtained through self-supervised pretraining. In Figure \ref{fig:loss_lr_pred}, we do not include DCASE as it is a very small dataset (total of $100$ recordings spread over $10$ classes) and already reached very high accuracy on both Kinetics400 and AudioSet.


\subsection{Comparison to the State-of-the-Arts} \label{sec:result}

\begin{table}[t]
\fontsize{9pt}{10pt}\selectfont
\centering

\resizebox{1\columnwidth}{!}{%
\begin{tabular}{lcccc}
\toprule
\bf Method & \bf {Compute} & \bf \specialcell{Backbone\\(\#Params (M))} & \bf U101 & \bf H51\\ \midrule\midrule

\multicolumn{3}{l}{Pretrained Dataset: Kinetics-Sound (Finetune input $32\!\times\!224^2$)} &&\\ \cmidrule{1-3}
CM-ACC\shortcite{cmacc} & 40 GPUs & 3D-ResNet18 (33.4)  & $77.2$ & $40.6$ \\ 
\textbf{CrissCross} & \textbf{4 GPUs} & R(2+1)D-18 (\textbf{15.4}) & $\mathbf{88.3}$ & $\mathbf{60.5}$ \\
\textcolor{fade}{Supervised~\shortcite{cmacc}} & - & \textcolor{fade}{3D-ResNet18 (33.4)} & \textcolor{fade}{$86.9$} & \textcolor{fade}{$53.1$} \\
\bottomrule

\multicolumn{3}{l}{Pretrained Dataset: Kinetics400 (Finetune input $8\!\times\!224^2$)} &&\\ \cmidrule{1-3}
XDC~\shortcite{xdc} & 64 GPUs & R(2+1)D-18 (31.5) & $74.2$ & $39.0$ \\ 
AVID~\shortcite{avid} & 64 GPUs & R(2+1)D-18 (15.4) & ${83.7}$ & ${49.5}$ \\
Robust-xID~\shortcite{ravid} & 8 GPUs & R(2+1)D-18 (15.4) & $81.9$ & $49.5$ \\
\textbf{CrissCross} & \textbf{8 GPUs} & R(2+1)D-18 (\textbf{15.4}) & $\mathbf{86.9}$ & $\mathbf{54.3}$ \\
\midrule
\multicolumn{3}{l}{Pretrained Dataset: Kinetics400 (Finetune input $32\!\times\!224^2$)} &&\\ \cmidrule{1-3}

SeLaVi~\shortcite{selavi} & 64 GPUs & R(2+1)D-18 (31.5) & $83.1$ & $47.1$ \\ 
XDC~\shortcite{xdc} & 64 GPUs & R(2+1)D-18 (31.5) & $86.8$ & $52.6$ \\ 
CM-ACC$^*$~\shortcite{cmacc}  & 40 GPUs & 3D-ResNet18 (33.4) & $90.2$ & $61.8$ \\ 
AVID~\shortcite{avid} & 64 GPUs & R(2+1)D-18 (15.4) & $87.5$ & ${60.8}$ \\ 
GDT~\shortcite{gdt} & 64 GPUs & R(2+1)D-18 (31.5) & ${90.9}$ & $62.3$ \\ 
CMAC~\shortcite{cmac} & 8 GPUs & R(2+1)D-18 (31.5) & $90.3$ & $61.1$ \\
Robust-xID~\shortcite{ravid} & 8 GPUs & R(2+1)D-18 (15.4) & $85.6$ & $55.0$ \\ 
\textbf{CrissCross} & \textbf{8 GPUs} & R(2+1)D-18 (\textbf{15.4}) & $\mathbf{91.5}$ & $\mathbf{64.7}$ \\ 
\textcolor{fade}{Supervised~\shortcite{gdt}} & - & \textcolor{fade}{R(2+1)D-18 (31.5)} & \textcolor{fade}{95.0} & \textcolor{fade}{74.0} \\ 
\bottomrule

\multicolumn{3}{l}{Pretrained Dataset: AudioSet (Finetune input $8\!\times\!224^2$)} &&\\ \cmidrule{1-3}

XDC~\shortcite{xdc} & 64 GPUs & R(2+1)D-18 (31.5) & $84.9$ & $48.8$ \\
AVID~\shortcite{avid} & 64 GPUs & R(2+1)D-18 (15.4) & ${88.6}$ & ${57.6}$ \\
\textbf{CrissCross} & \textbf{8 GPUs} & R(2+1)D-18 (\textbf{15.4}) & $\mathbf{89.4}$ & $\mathbf{58.3}$ \\
\midrule
\multicolumn{3}{l}{Pretrained Dataset: AudioSet (Finetune input $32\!\times\!224^2$)} &&\\ \cmidrule{1-3}
XDC~\shortcite{xdc} & 64 GPUs & R(2+1)D-18 (31.5) & $93.0$ & $63.7$ \\ 
MMV~\shortcite{mmv} & 32 TPUs & R(2+1)D-18 (31.5) & $91.5$ & $70.1$ \\ 
CM-ACC~\shortcite{cmacc} & 40 GPUs & R(2+1)D-18 (33.4) & $93.5$ & $67.2$ \\ 
BraVe$^{**}$~\shortcite{brave} & 16 TPUs & R(2+1)D-18 (31.5)  & $\mathbf{93.6}$ & $\mathbf{70.8}$ \\
AVID~\shortcite{avid} & 64 GPUs & R(2+1)D-18 (15.4) & $91.5$ & $64.7$ \\

\textbf{CrissCross} & \textbf{8 GPUs} & R(2+1)D-18 (\textbf{15.4}) & ${92.4}$ & ${67.4}$ \\ 
\textcolor{fade}{Supervised~\shortcite{brave}} & - & \textcolor{fade}{R(2+1)D-18 (31.5)} & \textcolor{fade}{96.8} & \textcolor{fade}{75.9} \\ 

\bottomrule

\multicolumn{5}{p{10cm}}{
$^*$ refers to 240K samples from Kinetics700.
$^{**}$ pretrained with very high temporal resolutions (2 views of $32~\&~128$ frames) compared to  others ($8/16/32$).
} \\

\end{tabular}%
}
\caption{{SOTA comparison on action recognition.} 
}
\label{tab:sota_action}
\end{table}

\begin{table}[tb]
    \fontsize{9pt}{10pt}\selectfont
    \setlength\tabcolsep{1pt}
    \centering
    
    \resizebox{0.4\textwidth}{!}{%
    \definecolor{rc1}{RGB}{235,235,235}
    \definecolor{rc2}{RGB}{255,255,255}
    \begin{tabular}{lcccccccc}
    \toprule
    \multirow{2}{*}{\bf Method} & \multicolumn{3}{c}{\bf UCF101} && \multicolumn{3}{c}{\bf HMDB51} \\ \cmidrule{2-4}\cmidrule{6-8}
    & R$@1$ & R$@5$ & R$@20$ & & R$@1$ & R$@5$ & R$@20$ \\ \midrule\midrule
    ST Order~\shortcite{storder}     & $25.7$ & $36.2$ & $49.2$ && - & - & - \\
    SpeedNet~\shortcite{speednet}    & $13.0$ & $28.1$ & $49.5$ && - & - & - \\
    Clip Order~\shortcite{cliporder} & $14.1$ & $30.3$ & $51.1$ && $7.6$ & $22.9$ & $48.8$ \\
    VCP~\shortcite{vcp}              & $18.6$ & $33.6$ & $53.5$ && $7.6$ & $24.4$ & $53.6$ \\
    VSP~\shortcite{vsp}              & $24.6$ & $41.9$ & $76.9$ && $10.3$ & $26.6$ & $54.6$ \\
    CoCLR~\shortcite{coclr}          & $55.9$ & $70.8$ & $82.5$ && $26.1$ & $45.8$ & $69.7$ \\
    SeLaVi~\shortcite{selavi}        & $52.0$ & $68.6$ & $84.5$ && $24.8$ & $47.6$ & $75.5$ \\
    Robust-xID~\shortcite{ravid}     & $60.9$ & $79.4$ & $90.8$ && $\mathbf{30.8}$ & $55.8$ & $79.7$ \\ 
    GDT~\shortcite{gdt}              & $57.4$ & $73.4$ & $88.1$ && $25.4$ & $51.4$ & $75.0$ \\
    \midrule
    \textbf{CrissCross}           & $\mathbf{63.8}$ & ${78.7}$ & ${89.9}$ && ${26.4}$ & ${50.5}$ & ${77.7}$ \\
    
    \bottomrule
    \end{tabular}%
    
    }
    \caption{{SOTA comparison on action retrieval.} 
    }
    \label{tab:sota-action_retrieval}
\end{table}

\begin{table}[!h]
    \centering
    \fontsize{9pt}{10pt}\selectfont
    \begin{tabular}{lcclcc}
    \toprule
    \multirow{2}{*}{\textbf{Method}} & \multicolumn{2}{c}{\textbf{ESC50}} && \multicolumn{2}{c}{\textbf{DCASE}} \\ \cmidrule{2-3} \cmidrule{5-6}
    & \textbf{K400} & \textbf{AS} & & \textbf{K400} & \textbf{AS} \\ \midrule\midrule
    AVTS~\shortcite{avts} & $76.7$ & $80.6$ &&  $91$ & $93$ \\ 
    XDC~\shortcite{xdc} & $78.0$ & $84.8$ &&  $91$ & $95$ \\ 
    AVID~\shortcite{avid} & $79.1$ & $89.1$ &&  $93$ & $96$ \\ 
    MMV~\shortcite{mmv}  & - & $85.6$ && -&- \\ 
    BraVe~\shortcite{brave} & - & $90.4$ &&-&- \\
    \textbf{CrissCross} & $\mathbf{86.8}$ & $\mathbf{90.5}$  && $\mathbf{96}$ & $\mathbf{97}$  \\
    \bottomrule
    \end{tabular}
    \caption{{SOTA comparison on sound classification.} 
    }
    \label{tab:sota_sound}
    
\end{table}

\subsubsection{Action Recognition.}
In line with \cite{xdc,selavi,avid,gdt,cmacc}, we benchmark CrissCross using UCF101 and HMDB51 on action recognition. For a fair comparison to earlier works, we adopt $2$ setups for finetuning, once with 8 frames, and the other with $32$ frames. In both these setups, we use a spatial resolution of $224^2$. We tune the model using the split-1 of both datasets and report the top-1 accuracy averaged over all the splits. We notice large variability in experimental setups in the literature in terms of different backbones (e.g., deeper ConvNets, Transformer-based architectures, etc.) \cite{elo,cvrl,stica}, pretraining inputs (e.g., the addition of optical flow or text in addition to audio-visual data, etc.) \cite{elo,cvrl,mmv}, and pretraining datasets, making it impractical to compare to all the prior works. Following the inclusion criteria of earlier works \cite{gdt,xdc,avid}, we compare CrissCross with methods that use similar backbones, inputs, and pretraining datasets.

The comparison of CrissCross with recent works is presented in Table \ref{tab:sota_action}. 
When pretrained with Kinetics400, CrissCross outperforms all the prior works by considerable margins on UCF101 and HMDB51 in both the fine-tuning setups. Moreover, CrissCross outperforms the current state-of-the-art AVID \cite{avid}, when pretrained on AudioSet and fine-tuned with $8$-frame inputs, on both the UCF101 and HMDB51. When fine-tuned with $32$-frame inputs, CrissCross achieves competitive results amongst the leading methods. We note that some of the prior works show slightly better performance compared to ours in some settings. We conjecture this to be due to the use of higher spatio-temporal resolution pretraining inputs in these models. E.g., BraVe \cite{brave} is pretrained with $2$ views of $32\!\times\!112^2$ and $128\!\times\!112^2$, and the input size for MMV \cite{mmv} and CM-ACC \cite{cmacc} are $32\!\times\!224^2$ and $16\!\times\!224^2$, respectively. In comparison, CrissCross is pretrained with visual inputs of size $8\!\times\!112^2$. However, we expect the performance of our model to improve further by using such higher resolutions, given the trend shown in \cite{brave}. 

In addition to the commonly used Kinectis400 and AudioSet, we further evaluate CrissCross while pretrained on the small-scale Kinetics-Sound. Here, we observe significant improvements compared to the current state-of-the-art CM-ACC \cite{cmacc} on both UCF101 ($88.3$ vs. $77.2$) and HMDB51 ($60.5$ vs. $40.6$). Additionally, CrissCross outperforms fully-supervised pretraining by $1.4\%$ and $7.4\%$ on UCF101 and HMDB51 respectively when both the fully-supervised and self-supervised methods are pretrained on Kinetics-Sound. To the best of our knowledge, this is the first time that self-supervision outperforms full-supervised pretraining on action recognition using the same small-scale pretraining dataset, showing that our method performs well on limited pretraining data.

\subsubsection{Action Retrieval.}
In addition to full finetuning, we also compare the performance of CrissCross in an unsupervised setup. Following prior works \cite{ravid,gdt,selavi}, we perform action retrieval using the split-1 of both UCF101 and HMDB51. The results are presented in Table \ref{tab:sota-action_retrieval} shows that CrissCross outperforms the current state-of-the-arts on UCF101 while achieving competitive results for HMDB51. 

\subsubsection{Sound Classification.}

We use two popular benchmarks ESC50 and DCASE to perform sound classification. We find large variability of experimental setups in the literature for evaluating audio representations. For instance, different backbones, input lengths, datasets, and evaluation protocols (linear evaluation, full-finetuning) have been used, making it impractical to compare to all the prior works. Following \cite{brave,mmv}, we perform linear evaluations using $5$-second inputs on ESC50 and $1$-second input for DCASE. As presented in Table \ref{tab:sota_sound}, CrissCross outperforms current state-of-the-art AVID \cite{avid} and BraVe \cite{brave} on ESC50, while pretrained on Kinetics400 and AudioSet respectively. Additionally, CrissCross sets new state-of-the-art by outperforming all the prior works on DCASE when pretrained on both Kinetics400 and AudioSet. 


\section{Summary} \label{sec:summary}
We propose a novel self-supervised framework to learn audio-visual representations by exploiting intra-modal, as well as, synchronous and \textit{asynchronous} cross-modal relationships. We conduct a thorough study investigating the major concepts of our framework. Our findings show that 
relaxation of cross-modal temporal synchronicity is beneficial for learning effective audio-visual representations. 
These representations can then be used for a variety of downstream tasks including action recognition, sound classification, and action retrieval.

\subsubsection*{Acknowledgments}
We are grateful to the Bank of Montreal and Mitacs for funding this research. We are thankful to SciNet HPC Consortium for helping with the computation resources.

\bibliography{refs}

\begin{thebibliography}{68}
\providecommand{\natexlab}[1]{#1}

\bibitem[{Ahsan, Madhok, and Essa(2019)}]{ahsan2019video}
Ahsan, U.; Madhok, R.; and Essa, I. 2019.
\newblock Video jigsaw: Unsupervised learning of spatiotemporal context for
  video action recognition.
\newblock In \emph{WACV}, 179--189.

\bibitem[{Alayrac et~al.(2020)Alayrac, Recasens, Schneider, Arandjelovic,
  Ramapuram, De~Fauw, Smaira, Dieleman, and Zisserman}]{mmv}
Alayrac, J.-B.; Recasens, A.; Schneider, R.; Arandjelovic, R.; Ramapuram, J.;
  De~Fauw, J.; Smaira, L.; Dieleman, S.; and Zisserman, A. 2020.
\newblock Self-Supervised MultiModal Versatile Networks.
\newblock \emph{NeurIPS}, 2(6): 7.

\bibitem[{Alwassel et~al.(2020)Alwassel, Mahajan, Korbar, Torresani, Ghanem,
  and Tran}]{xdc}
Alwassel, H.; Mahajan, D.; Korbar, B.; Torresani, L.; Ghanem, B.; and Tran, D.
  2020.
\newblock Self-Supervised Learning by Cross-Modal Audio-Video Clustering.
\newblock \emph{NeruIPS}, 33.

\bibitem[{Arandjelovic and Zisserman(2017)}]{l3-kineticssound}
Arandjelovic, R.; and Zisserman, A. 2017.
\newblock Look, listen and learn.
\newblock In \emph{ICCV}, 609--617.

\bibitem[{Asano et~al.(2020)Asano, Patrick, Rupprecht, and Vedaldi}]{selavi}
Asano, Y.~M.; Patrick, M.; Rupprecht, C.; and Vedaldi, A. 2020.
\newblock Labelling unlabelled videos from scratch with multi-modal
  self-supervision.
\newblock In \emph{NeurIPS}.

\bibitem[{Babaeizadeh et~al.(2018)Babaeizadeh, Finn, Erhan, Campbell, and
  Levine}]{babaeizadeh2017stochastic}
Babaeizadeh, M.; Finn, C.; Erhan, D.; Campbell, R.~H.; and Levine, S. 2018.
\newblock Stochastic Variational Video Prediction.
\newblock In \emph{ICLR}.

\bibitem[{Bardes, Ponce, and LeCun(2021)}]{vicreg}
Bardes, A.; Ponce, J.; and LeCun, Y. 2021.
\newblock Vicreg: Variance-invariance-covariance regularization for
  self-supervised learning.
\newblock \emph{arXiv preprint arXiv:2105.04906}.

\bibitem[{Benaim et~al.(2020)Benaim, Ephrat, Lang, Mosseri, Freeman,
  Rubinstein, Irani, and Dekel}]{speednet}
Benaim, S.; Ephrat, A.; Lang, O.; Mosseri, I.; Freeman, W.~T.; Rubinstein, M.;
  Irani, M.; and Dekel, T. 2020.
\newblock Speednet: Learning the speediness in videos.
\newblock In \emph{CVPR}, 9922--9931.

\bibitem[{Bromley et~al.(1993)Bromley, Guyon, LeCun, S{\"a}ckinger, and
  Shah}]{bromley1993signature}
Bromley, J.; Guyon, I.; LeCun, Y.; S{\"a}ckinger, E.; and Shah, R. 1993.
\newblock Signature verification using a" siamese" time delay neural network.
\newblock \emph{Advances in neural information processing systems}, 6.

\bibitem[{Buchler, Brattoli, and Ommer(2018)}]{storder}
Buchler, U.; Brattoli, B.; and Ommer, B. 2018.
\newblock Improving spatiotemporal self-supervision by deep reinforcement
  learning.
\newblock In \emph{ECCV}.

\bibitem[{Caron et~al.(2018)Caron, Bojanowski, Joulin, and Douze}]{deepcluster}
Caron, M.; Bojanowski, P.; Joulin, A.; and Douze, M. 2018.
\newblock Deep clustering for unsupervised learning of visual features.
\newblock In \emph{ECCV}, 132--149.

\bibitem[{Caron et~al.(2020)Caron, Misra, Mairal, Goyal, Bojanowski, and
  Joulin}]{swav}
Caron, M.; Misra, I.; Mairal, J.; Goyal, P.; Bojanowski, P.; and Joulin, A.
  2020.
\newblock Unsupervised Learning of Visual Features by Contrasting Cluster
  Assignments.
\newblock In \emph{NeurIPS}.

\bibitem[{Chen et~al.(2020)Chen, Kornblith, Norouzi, and Hinton}]{simclr}
Chen, T.; Kornblith, S.; Norouzi, M.; and Hinton, G. 2020.
\newblock A simple framework for contrastive learning of visual
  representations.
\newblock In \emph{ICML}, 1597--1607.

\bibitem[{Chen and He(2021)}]{simsiam}
Chen, X.; and He, K. 2021.
\newblock Exploring simple siamese representation learning.
\newblock In \emph{CVPR}, 15750--15758.

\bibitem[{Cho et~al.(2020)Cho, Kim, Chang, and Hwang}]{vsp}
Cho, H.; Kim, T.; Chang, H.~J.; and Hwang, W. 2020.
\newblock Self-Supervised Spatio-Temporal Representation Learning Using
  Variable Playback Speed Prediction.
\newblock \emph{arXiv preprint arXiv:2003.02692}.

\bibitem[{DeVries and Taylor(2017)}]{cutout}
DeVries, T.; and Taylor, G.~W. 2017.
\newblock Improved regularization of convolutional neural networks with cutout.
\newblock \emph{arXiv preprint arXiv:1708.04552}.

\bibitem[{Finn, Goodfellow, and Levine(2016)}]{finn2016unsupervised}
Finn, C.; Goodfellow, I.; and Levine, S. 2016.
\newblock Unsupervised learning for physical interaction through video
  prediction.
\newblock \emph{NeurIPS}, 29: 64--72.

\bibitem[{Gemmeke et~al.(2017)Gemmeke, Ellis, Freedman, Jansen, Lawrence,
  Moore, Plakal, and Ritter}]{audioset}
Gemmeke, J.~F.; Ellis, D.~P.; Freedman, D.; Jansen, A.; Lawrence, W.; Moore,
  R.~C.; Plakal, M.; and Ritter, M. 2017.
\newblock Audio set: An ontology and human-labeled dataset for audio events.
\newblock In \emph{ICASSP}, 776--780.

\bibitem[{Ghadiyaram, Tran, and Mahajan(2019)}]{ig65m}
Ghadiyaram, D.; Tran, D.; and Mahajan, D. 2019.
\newblock Large-Scale Weakly-Supervised Pre-Training for Video Action
  Recognition.
\newblock In \emph{CVPR}, 12038--12047.

\bibitem[{Grill et~al.(2020)Grill, Strub, Altch{\'e}, Tallec, Richemond,
  Buchatskaya, Doersch, Pires, Guo, Azar et~al.}]{byol}
Grill, J.-B.; Strub, F.; Altch{\'e}, F.; Tallec, C.; Richemond, P.;
  Buchatskaya, E.; Doersch, C.; Pires, B.; Guo, Z.; Azar, M.; et~al. 2020.
\newblock Bootstrap Your Own Latent: A new approach to self-supervised
  learning.
\newblock In \emph{NeurIPS}.

\bibitem[{Han, Xie, and Zisserman(2020)}]{coclr}
Han, T.; Xie, W.; and Zisserman, A. 2020.
\newblock Self-supervised Co-training for Video Representation Learning.
\newblock In \emph{NeurIPS}.

\bibitem[{He et~al.(2016)He, Zhang, Ren, and Sun}]{resnet}
He, K.; Zhang, X.; Ren, S.; and Sun, J. 2016.
\newblock Deep residual learning for image recognition.
\newblock In \emph{CVPR}, 770--778.

\bibitem[{Jing et~al.(2018)Jing, Yang, Liu, and Tian}]{rotnet3d}
Jing, L.; Yang, X.; Liu, J.; and Tian, Y. 2018.
\newblock Self-supervised spatiotemporal feature learning via video rotation
  prediction.
\newblock \emph{arXiv preprint arXiv:1811.11387}.

\bibitem[{Kay et~al.(2017)Kay, Carreira, Simonyan, Zhang, Hillier,
  Vijayanarasimhan, Viola, Green, Back, Natsev et~al.}]{kinetics400}
Kay, W.; Carreira, J.; Simonyan, K.; Zhang, B.; Hillier, C.; Vijayanarasimhan,
  S.; Viola, F.; Green, T.; Back, T.; Natsev, P.; et~al. 2017.
\newblock The kinetics human action video dataset.
\newblock \emph{arXiv preprint arXiv:1705.06950}.

\bibitem[{Khare, Parthasarathy, and Sundaram(2021)}]{khare2021self}
Khare, A.; Parthasarathy, S.; and Sundaram, S. 2021.
\newblock Self-Supervised learning with cross-modal transformers for emotion
  recognition.
\newblock In \emph{SLT}, 381--388.

\bibitem[{Kim, Cho, and Kweon(2019)}]{stc}
Kim, D.; Cho, D.; and Kweon, I.~S. 2019.
\newblock Self-supervised video representation learning with space-time cubic
  puzzles.
\newblock In \emph{AAAI}, volume~33, 8545--8552.

\bibitem[{Kingma and Ba(2015)}]{adam}
Kingma, D.~P.; and Ba, J. 2015.
\newblock Adam: A Method for Stochastic Optimization.
\newblock In \emph{ICLR}.

\bibitem[{Korbar, Tran, and Torresani(2018)}]{avts}
Korbar, B.; Tran, D.; and Torresani, L. 2018.
\newblock Cooperative learning of audio and video models from self-supervised
  synchronization.
\newblock In \emph{NeruIPS}, 7774--7785.

\bibitem[{Kuehne et~al.(2011)Kuehne, Jhuang, Garrote, Poggio, and Serre}]{hmdb}
Kuehne, H.; Jhuang, H.; Garrote, E.; Poggio, T.; and Serre, T. 2011.
\newblock HMDB: a large video database for human motion recognition.
\newblock In \emph{ICCV}, 2556--2563.

\bibitem[{Lee et~al.(2017)Lee, Huang, Singh, and Yang}]{opn}
Lee, H.-Y.; Huang, J.-B.; Singh, M.; and Yang, M.-H. 2017.
\newblock Unsupervised representation learning by sorting sequences.
\newblock In \emph{CVPR}.

\bibitem[{Liang et~al.(2017)Liang, Lee, Dai, and Xing}]{liang2017dual}
Liang, X.; Lee, L.; Dai, W.; and Xing, E.~P. 2017.
\newblock Dual motion gan for future-flow embedded video prediction.
\newblock In \emph{ICCV}, 1744--1752.

\bibitem[{Loshchilov and Hutter(2017)}]{cosine_lrs}
Loshchilov, I.; and Hutter, F. 2017.
\newblock Sgdr: Stochastic gradient descent with warm restarts.
\newblock In \emph{ICLR}.

\bibitem[{Luo et~al.(2020)Luo, Liu, Zhou, Yang, Ma, Ye, and Wang}]{vcp}
Luo, D.; Liu, C.; Zhou, Y.; Yang, D.; Ma, C.; Ye, Q.; and Wang, W. 2020.
\newblock Video cloze procedure for self-supervised spatio-temporal learning.
\newblock In \emph{AAAI}.

\bibitem[{Ma et~al.(2020)Ma, Zeng, McDuff, and Song}]{cmacc}
Ma, S.; Zeng, Z.; McDuff, D.; and Song, Y. 2020.
\newblock Active Contrastive Learning of Audio-Visual Video Representations.
\newblock In \emph{ICLR}.

\bibitem[{Mathieu, Couprie, and LeCun(2016)}]{mathieu2015deep}
Mathieu, M.; Couprie, C.; and LeCun, Y. 2016.
\newblock Deep multi-scale video prediction beyond mean square error.
\newblock In \emph{ICLR}.

\bibitem[{McFee et~al.(2015)McFee, Raffel, Liang, Ellis, McVicar, Battenberg,
  and Nieto}]{librosa}
McFee, B.; Raffel, C.; Liang, D.; Ellis, D.~P.; McVicar, M.; Battenberg, E.;
  and Nieto, O. 2015.
\newblock librosa: Audio and music signal analysis in python.
\newblock In \emph{Python in Science Conference}, volume~8, 18--25.

\bibitem[{Micikevicius et~al.(2018)Micikevicius, Narang, Alben, Diamos, Elsen,
  Garcia, Ginsburg, Houston, Kuchaiev, Venkatesh et~al.}]{amp}
Micikevicius, P.; Narang, S.; Alben, J.; Diamos, G.; Elsen, E.; Garcia, D.;
  Ginsburg, B.; Houston, M.; Kuchaiev, O.; Venkatesh, G.; et~al. 2018.
\newblock Mixed Precision Training.
\newblock In \emph{ICLR}.

\bibitem[{Min et~al.(2021)Min, Dai, Xie, Gan, Zhang, and Wang}]{cmac}
Min, S.; Dai, Q.; Xie, H.; Gan, C.; Zhang, Y.; and Wang, J. 2021.
\newblock Cross-Modal Attention Consistency for Video-Audio Unsupervised
  Learning.
\newblock \emph{arXiv preprint arXiv:2106.06939}.

\bibitem[{Misra and Maaten(2020)}]{pirl}
Misra, I.; and Maaten, L. v.~d. 2020.
\newblock Self-supervised learning of pretext-invariant representations.
\newblock In \emph{CVPR}, 6707--6717.

\bibitem[{Misra, Zitnick, and Hebert(2016)}]{shuffle-learn}
Misra, I.; Zitnick, C.~L.; and Hebert, M. 2016.
\newblock Shuffle and learn: unsupervised learning using temporal order
  verification.
\newblock In \emph{ECCV}, 527--544.

\bibitem[{Morgado, Misra, and Vasconcelos(2021)}]{ravid}
Morgado, P.; Misra, I.; and Vasconcelos, N. 2021.
\newblock Robust Audio-Visual Instance Discrimination.
\newblock In \emph{CVPR}, 12934--12945.

\bibitem[{Morgado, Vasconcelos, and Misra(2021)}]{avid}
Morgado, P.; Vasconcelos, N.; and Misra, I. 2021.
\newblock Audio-visual instance discrimination with cross-modal agreement.
\newblock In \emph{CVPR}, 12475--12486.

\bibitem[{Niizumi et~al.(2021)Niizumi, Takeuchi, Ohishi, Harada, and
  Kashino}]{byol_audio}
Niizumi, D.; Takeuchi, D.; Ohishi, Y.; Harada, N.; and Kashino, K. 2021.
\newblock BYOL for Audio: Self-Supervised Learning for General-Purpose Audio
  Representation.
\newblock \emph{arXiv preprint arXiv:2103.06695}.

\bibitem[{Omeiza et~al.(2019)Omeiza, Speakman, Cintas, and
  Weldermariam}]{omeiza2019smooth}
Omeiza, D.; Speakman, S.; Cintas, C.; and Weldermariam, K. 2019.
\newblock Smooth grad-cam++: An enhanced inference level visualization
  technique for deep convolutional neural network models.
\newblock \emph{arXiv preprint arXiv:1908.01224}.

\bibitem[{Park et~al.(2019)Park, Chan, Zhang, Chiu, Zoph, Cubuk, and
  Le}]{specaug}
Park, D.~S.; Chan, W.; Zhang, Y.; Chiu, C.-C.; Zoph, B.; Cubuk, E.~D.; and Le,
  Q.~V. 2019.
\newblock Specaugment: A simple data augmentation method for automatic speech
  recognition.
\newblock \emph{arXiv preprint arXiv:1904.08779}.

\bibitem[{Paszke et~al.(2019)Paszke, Gross, Massa, Lerer, Bradbury, Chanan,
  Killeen, Lin, Gimelshein, Antiga et~al.}]{Pytorch}
Paszke, A.; Gross, S.; Massa, F.; Lerer, A.; Bradbury, J.; Chanan, G.; Killeen,
  T.; Lin, Z.; Gimelshein, N.; Antiga, L.; et~al. 2019.
\newblock Pytorch: An imperative style, high-performance deep learning library.
\newblock \emph{NeurIPS}, 32: 8026--8037.

\bibitem[{Patrick et~al.(2021{\natexlab{a}})Patrick, Asano, Kuznetsova, Fong,
  Henriques, Zweig, and Vedaldi}]{gdt}
Patrick, M.; Asano, Y.~M.; Kuznetsova, P.; Fong, R.; Henriques, J.~F.; Zweig,
  G.; and Vedaldi, A. 2021{\natexlab{a}}.
\newblock On compositions of transformations in contrastive self-supervised
  learning.
\newblock In \emph{Proceedings of the IEEE/CVF International Conference on
  Computer Vision}, 9577--9587.

\bibitem[{Patrick et~al.(2021{\natexlab{b}})Patrick, Huang, Misra, Metze,
  Vedaldi, Asano, and Henriques}]{stica}
Patrick, M.; Huang, P.-Y.; Misra, I.; Metze, F.; Vedaldi, A.; Asano, Y.~M.; and
  Henriques, J.~F. 2021{\natexlab{b}}.
\newblock Space-Time Crop \& Attend: Improving Cross-modal Video Representation
  Learning.
\newblock In \emph{ICCV}, 10560--10572.

\bibitem[{Piczak(2015)}]{esc}
Piczak, K.~J. 2015.
\newblock {ESC}: {Dataset} for {Environmental Sound Classification}.
\newblock In \emph{ACM Conference on Multimedia}, 1015--1018. {}.

\bibitem[{Piergiovanni, Angelova, and Ryoo(2020)}]{elo}
Piergiovanni, A.; Angelova, A.; and Ryoo, M.~S. 2020.
\newblock Evolving losses for unsupervised video representation learning.
\newblock In \emph{CVPR}, 133--142.

\bibitem[{Qian et~al.(2021)Qian, Meng, Gong, Yang, Wang, Belongie, and
  Cui}]{cvrl}
Qian, R.; Meng, T.; Gong, B.; Yang, M.-H.; Wang, H.; Belongie, S.; and Cui, Y.
  2021.
\newblock Spatiotemporal contrastive video representation learning.
\newblock In \emph{CVPR}, 6964--6974.

\bibitem[{Recasens et~al.(2021)Recasens, Luc, Alayrac, Wang, Strub, Tallec,
  Malinowski, Patraucean, Altch{\'e}, Valko et~al.}]{brave}
Recasens, A.; Luc, P.; Alayrac, J.-B.; Wang, L.; Strub, F.; Tallec, C.;
  Malinowski, M.; Patraucean, V.; Altch{\'e}, F.; Valko, M.; et~al. 2021.
\newblock Broaden Your Views for Self-Supervised Video Learning.
\newblock \emph{arXiv preprint arXiv:2103.16559}.

\bibitem[{Reda et~al.(2018)Reda, Liu, Shih, Kirby, Barker, Tarjan, Tao, and
  Catanzaro}]{reda2018sdc}
Reda, F.~A.; Liu, G.; Shih, K.~J.; Kirby, R.; Barker, J.; Tarjan, D.; Tao, A.;
  and Catanzaro, B. 2018.
\newblock Sdc-net: Video prediction using spatially-displaced convolution.
\newblock In \emph{ECCV}, 718--733.

\bibitem[{Saito, Matsumoto, and Saito(2017)}]{saito2017temporal}
Saito, M.; Matsumoto, E.; and Saito, S. 2017.
\newblock Temporal generative adversarial nets with singular value clipping.
\newblock In \emph{ICCV}, 2830--2839.

\bibitem[{Sarkar and Etemad(2020{\natexlab{a}})}]{sarkar-ssl-tafc}
Sarkar, P.; and Etemad, A. 2020{\natexlab{a}}.
\newblock Self-supervised ECG representation learning for emotion recognition.
\newblock \emph{IEEE Transactions on Affective Computing}.

\bibitem[{Sarkar and Etemad(2020{\natexlab{b}})}]{sarkar-ssl-icassp}
Sarkar, P.; and Etemad, A. 2020{\natexlab{b}}.
\newblock Self-supervised learning for ecg-based emotion recognition.
\newblock In \emph{ICASSP}, 3217--3221.

\bibitem[{Sarkar et~al.(2020)Sarkar, Lobmaier, Fabre, Berg, Mueller, Frasch,
  Antonelli, and Etemad}]{sarkar-ssl2}
Sarkar, P.; Lobmaier, S.; Fabre, B.; Berg, G.; Mueller, A.; Frasch, M.~G.;
  Antonelli, M.~C.; and Etemad, A. 2020.
\newblock Detection of Maternal and Fetal Stress from ECG with Self-supervised
  Representation Learning.
\newblock \emph{arXiv e-prints}, arXiv--2011.

\bibitem[{Siriwardhana et~al.(2020)Siriwardhana, Kaluarachchi, Billinghurst,
  and Nanayakkara}]{siriwardhana2020multimodal}
Siriwardhana, S.; Kaluarachchi, T.; Billinghurst, M.; and Nanayakkara, S. 2020.
\newblock Multimodal Emotion Recognition With Transformer-Based Self Supervised
  Feature Fusion.
\newblock \emph{IEEE Access}, 8: 176274--176285.

\bibitem[{Soomro, Zamir, and Shah(2012)}]{ucf101}
Soomro, K.; Zamir, A.~R.; and Shah, M. 2012.
\newblock UCF101: A dataset of 101 human actions classes from videos in the
  wild.
\newblock \emph{arXiv preprint arXiv:1212.0402}.

\bibitem[{Stowell et~al.(2015)Stowell, Giannoulis, Benetos, Lagrange, and
  Plumbley}]{dcase}
Stowell, D.; Giannoulis, D.; Benetos, E.; Lagrange, M.; and Plumbley, M.~D.
  2015.
\newblock Detection and classification of acoustic scenes and events.
\newblock \emph{IEEE Transactions on Multimedia}, 17(10): 1733--1746.

\bibitem[{Tran et~al.(2016)Tran, Bourdev, Fergus, Torresani, and
  Paluri}]{tran2016deep}
Tran, D.; Bourdev, L.; Fergus, R.; Torresani, L.; and Paluri, M. 2016.
\newblock Deep end2end voxel2voxel prediction.
\newblock In \emph{CVPRW}, 17--24.

\bibitem[{Tran et~al.(2018)Tran, Wang, Torresani, Ray, LeCun, and
  Paluri}]{r2plus1d}
Tran, D.; Wang, H.; Torresani, L.; Ray, J.; LeCun, Y.; and Paluri, M. 2018.
\newblock A closer look at spatiotemporal convolutions for action recognition.
\newblock In \emph{CVPR}, 6450--6459.

\bibitem[{Tulyakov et~al.(2018)Tulyakov, Liu, Yang, and
  Kautz}]{tulyakov2017mocogan}
Tulyakov, S.; Liu, M.-Y.; Yang, X.; and Kautz, J. 2018.
\newblock {MoCoGAN}: Decomposing motion and content for video generation.
\newblock In \emph{CVPR}, 1526--1535.

\bibitem[{Van~Gansbeke et~al.(2020)Van~Gansbeke, Vandenhende, Georgoulis,
  Proesmans, and Van~Gool}]{scan}
Van~Gansbeke, W.; Vandenhende, S.; Georgoulis, S.; Proesmans, M.; and Van~Gool,
  L. 2020.
\newblock Scan: Learning to classify images without labels.
\newblock In \emph{ECCV}, 268--285.

\bibitem[{Vondrick, Pirsiavash, and Torralba(2016)}]{vondrick2016generating}
Vondrick, C.; Pirsiavash, H.; and Torralba, A. 2016.
\newblock Generating videos with scene dynamics.
\newblock \emph{NeurIPS}, 29: 613--621.

\bibitem[{Wang et~al.(2021)Wang, Jiao, Bao, He, Liu, and Liu}]{wang2021self}
Wang, J.; Jiao, J.; Bao, L.; He, S.; Liu, W.; and Liu, Y.-H. 2021.
\newblock Self-supervised Video Representation Learning by Uncovering
  Spatio-temporal Statistics.
\newblock \emph{PAMI}.

\bibitem[{Xu et~al.(2019)Xu, Xiao, Zhao, Shao, Xie, and Zhuang}]{cliporder}
Xu, D.; Xiao, J.; Zhao, Z.; Shao, J.; Xie, D.; and Zhuang, Y. 2019.
\newblock Self-supervised spatiotemporal learning via video clip order
  prediction.
\newblock In \emph{CVPR}, 10334--10343.

\bibitem[{You, Gitman, and Ginsburg(2017)}]{larc}
You, Y.; Gitman, I.; and Ginsburg, B. 2017.
\newblock Large batch training of convolutional networks.
\newblock \emph{arXiv preprint arXiv:1708.03888}.

\end{thebibliography}

\clearpage
\appendix
\begin{center}
\LARGE
\textbf{Supplementary Material}
\end{center}

\setcounter{table}{0}
\setcounter{figure}{0}
\renewcommand{\thetable}{S\arabic{table}}
\renewcommand\thefigure{S\arabic{figure}}


The organization of the supplementary material is as follows:
\begin{itemize}[noitemsep,nolistsep]
    \item Appendix \ref{supsec:algo}: Pseudocode;
    \item Appendix \ref{supsec:qual_anal}: Qualitative Analysis;
    \item Appendix \ref{supsec:datasets}: Datasets;
    \item Appendix \ref{supsec:augs}: Data Augmentations;
    \item Appendix \ref{supsec:eval_proto}: Evaluation Protocols;
    \item Appendix \ref{supsec:hyperparams}: Hyperparameters;
    \item Appendix \ref{supsec:arch}: Architectures;
    \item Appendix \ref{supsec:limitations}: Limitations;
    \item Appendix \ref{supsec:borader_impact}: Broader Impact.
\end{itemize}


\section{Pseudocode} \label{supsec:algo}
We present the pseudocode of our proposed CrissCross framework in Algorithm \ref{algo:crisscross}. 

\begin{algorithm}[!h]
\caption{CrissCross pseudocode (PyTorch style).}
\label{algo:crisscross}
\lstset{
  numbers=none,
  backgroundcolor=\color{white},
  basicstyle=\fontsize{7.5pt}{7.5pt}\ttfamily\bfseries\selectfont,
  columns=fullflexible,
  breaklines=true,
  captionpos=b,
  commentstyle=\fontsize{7.5pt}{7.5pt}\color{codeblue},
  keywordstyle=\fontsize{7.5pt}{7.5pt}\color{codekw},
}

\begin{lstlisting}[language=python]
# fv: visual encoder (backbone+projection mlp)
# fa: audio encoder (backbone+projection mlp)
# hv: visual predictor head (prediction mlp)
# ha: audio predictor head (prediction mlp)
# D: loss function, following Eqn. 1

def forward(v1, v2, a1, a2):
    """
    v1,V2: minibatch of augmented visual samples
    a1,a2: minibatch of augmented audio samples
    """
    
    # visual
    zv1, zv2 = fv(v1), fv(v2)       # visual embeddings
    pv1, pv2 = hv(zv1), hv(zv2)     # predictor output
    
    # audio
    za1, za2 = fa(a1), fa(a2)       # audio embeddings
    pa1, pa2 = ha(za1), ha(za2)     # predictor output
    
    # loss calculation
    
    # intra-modal loss, following Eqn. 2
    L_intra = D(pv1, zv2)/2 + D(pv2, zv1)/2 + \
              D(pa1, za2)/2 + D(pa2, za1)/2
    
    # synchronous cross-modal loss, following Eqn. 3
    L_sync = (D(pv1, za1)/2 + D(pa1, zv1)/2 + Lv2a2 +\
              D(pv2, za2)/2 + D(pa2, zv2)/2)/2
    
    # asynchronous cross-modal loss, following Eqn. 4
    L_async = (D(pv1, za2)/2 + D(pa2, zv1)/2 +\
               D(pa1, zv2)/2 + D(pv2, za1)/2)/2
    
    # total loss, following Eqn. 5
    L_CrissCross = (L_async + L_sync + L_intra)/3
    
    return L_CrissCross
\end{lstlisting}
\end{algorithm}

\section{Qualitative Analysis} \label{supsec:qual_anal}
\begin{figure}[tb]
    \centering
    \resizebox{\linewidth}{!}{
    \begin{tabular}{rccccc}
    \textbf{\tiny{~~~~Query}} & \multicolumn{5}{c}{\textbf{\tiny{Neighborhoods}}} \\
    \multicolumn{6}{c}{\includegraphics[width=0.45\textwidth]{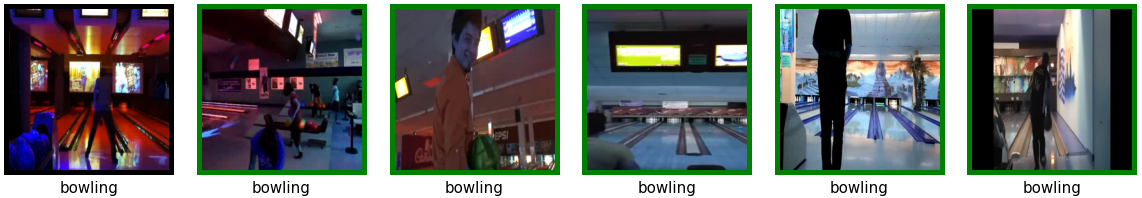}} \\
    \multicolumn{6}{c}{\includegraphics[width=0.45\textwidth]{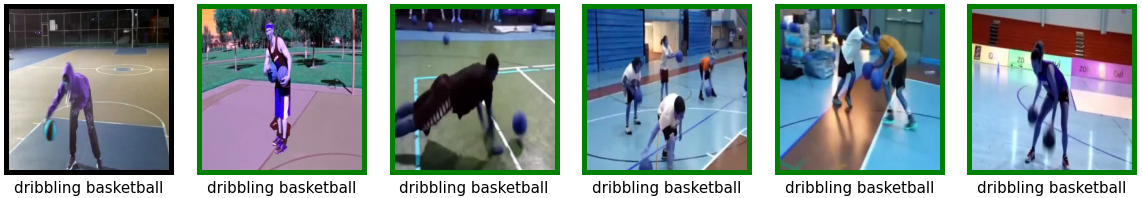}} \\
    \multicolumn{6}{c}{\includegraphics[width=0.45\textwidth]{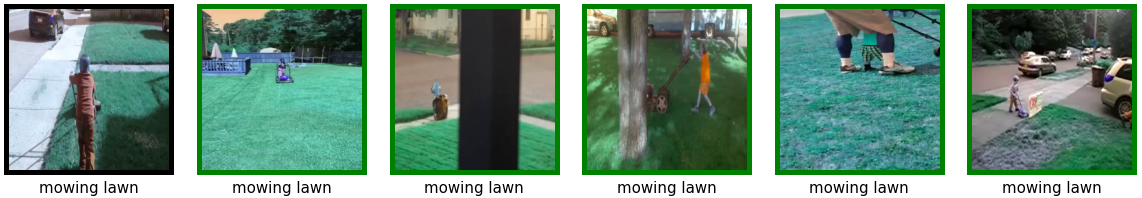}} \\
    \multicolumn{6}{c}{\includegraphics[width=0.45\textwidth]{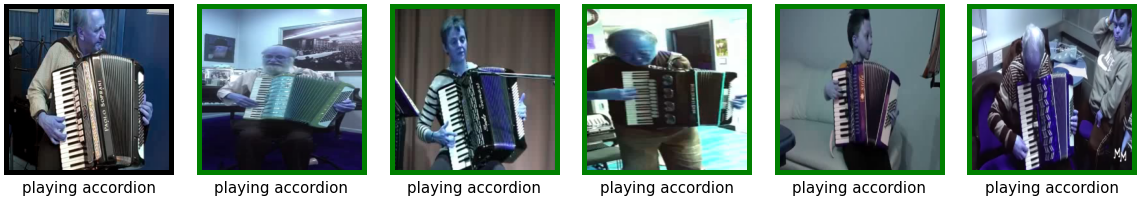}} \\
    \multicolumn{6}{c}{\includegraphics[width=0.45\textwidth]{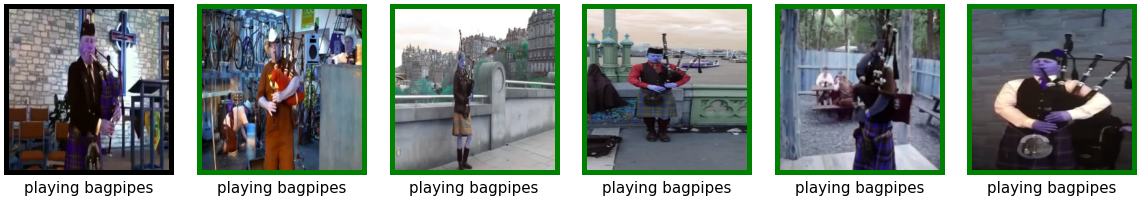}} \\
    \multicolumn{6}{c}{\includegraphics[width=0.45\textwidth]{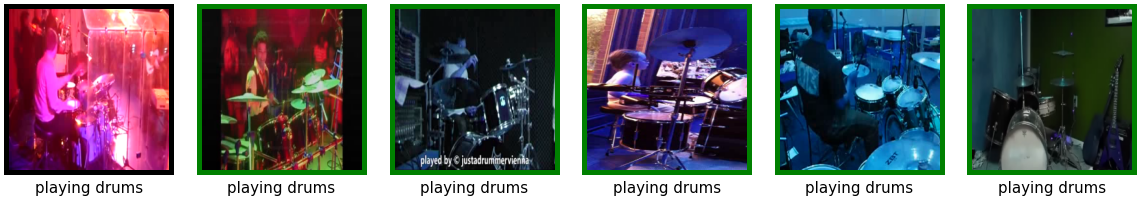}} \\
    \multicolumn{6}{c}{\includegraphics[width=0.45\textwidth]{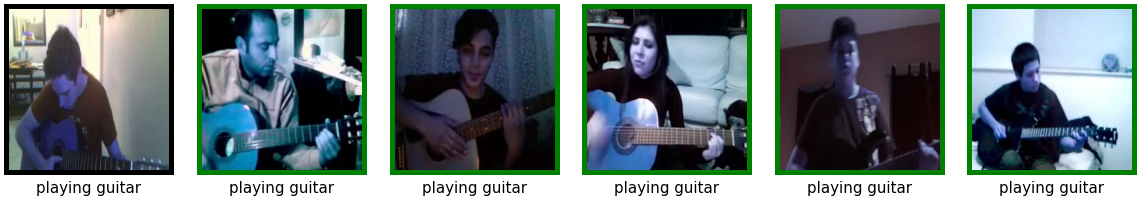}} \\
    \multicolumn{6}{c}{\includegraphics[width=0.45\textwidth]{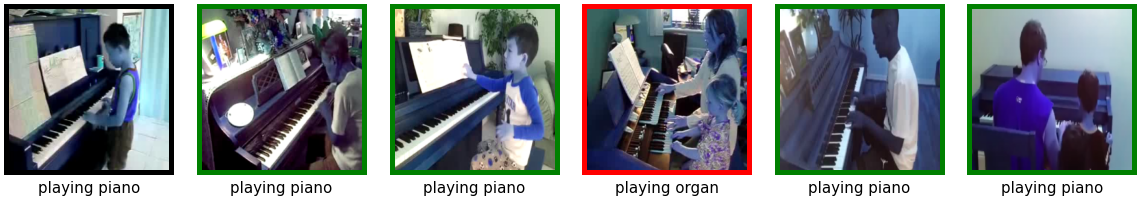}} \\
    \multicolumn{6}{c}{\includegraphics[width=0.45\textwidth]{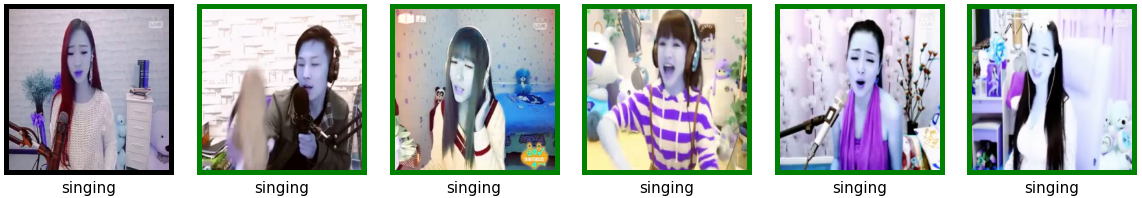}} \\

    \end{tabular}
    }
    \caption{{We present a few randomly selected samples of \textbf{video-to-video} retrieval.} Here, the frames with {black} borders represent the query, and the next $5$ frames represent the top-5 neighborhoods. The correct retrievals are marked with {{green}}, while the wrong ones are marked with {{red}}. 
    }
    \label{fig:ret_v2v}
\end{figure}

\begin{figure}[tb]
    \centering
    \resizebox{\linewidth}{!}{
    \begin{tabular}{rccccc}
    \textbf{\tiny{~~~~Query}} & \multicolumn{5}{c}{\textbf{\tiny{Neighborhoods}}} \\
    \multicolumn{6}{c}{\includegraphics[width=0.45\textwidth]{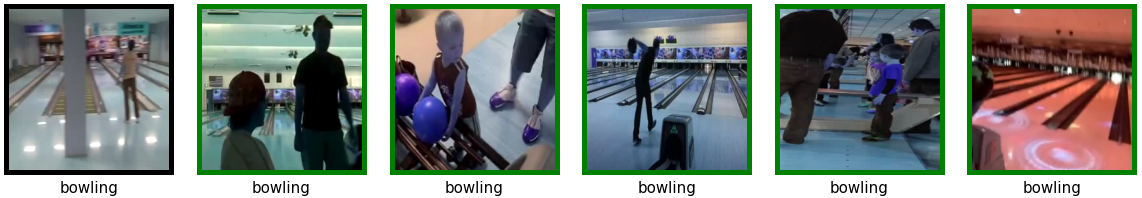}} \\
    \multicolumn{6}{c}{\includegraphics[width=0.45\textwidth]{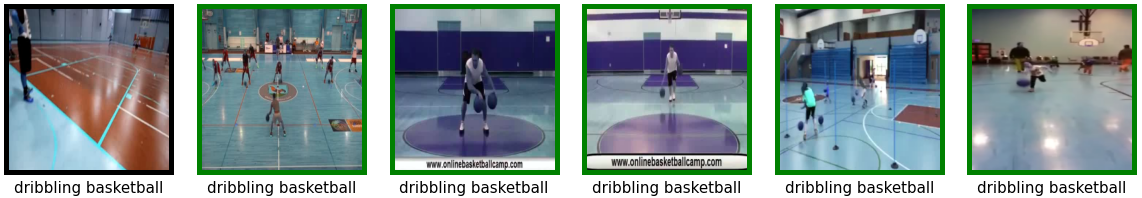}} \\
    \multicolumn{6}{c}{\includegraphics[width=0.45\textwidth]{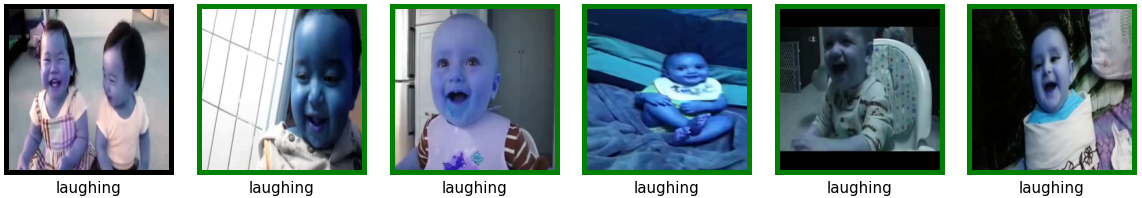}} \\
    \multicolumn{6}{c}{\includegraphics[width=0.45\textwidth]{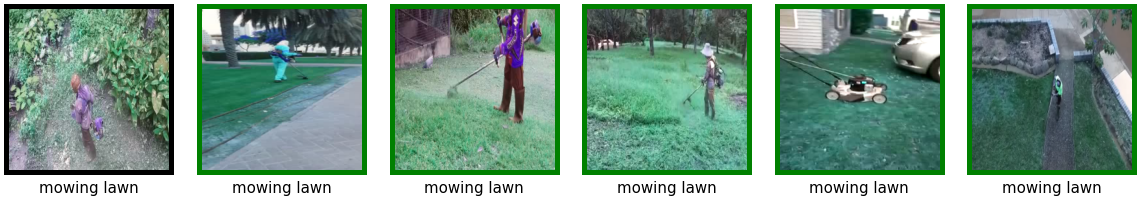}} \\
    \multicolumn{6}{c}{\includegraphics[width=0.45\textwidth]{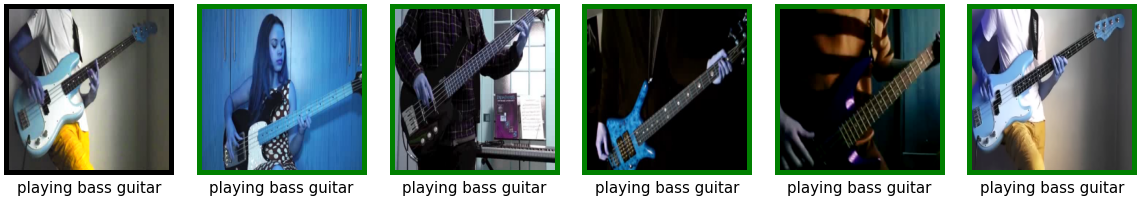}} \\
    \multicolumn{6}{c}{\includegraphics[width=0.45\textwidth]{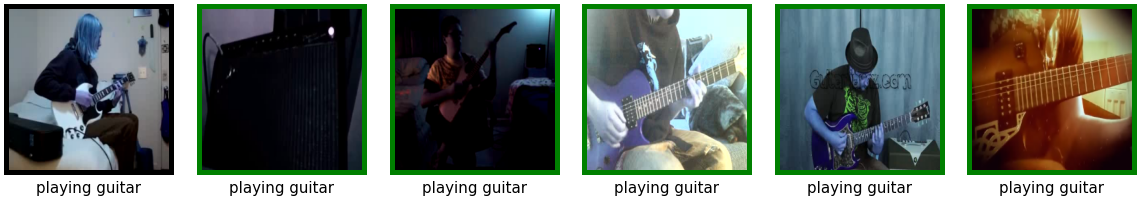}} \\
    \multicolumn{6}{c}{\includegraphics[width=0.45\textwidth]{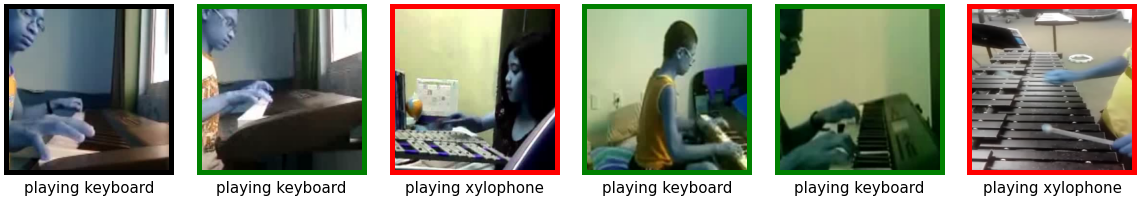}} \\
    \multicolumn{6}{c}{\includegraphics[width=0.45\textwidth]{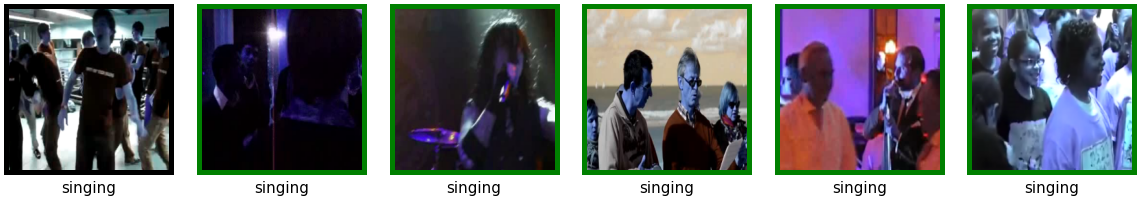}} \\
    \multicolumn{6}{c}{\includegraphics[width=0.45\textwidth]{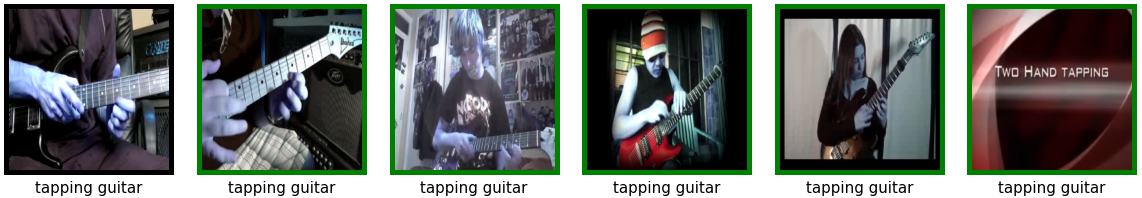}} \\

    \end{tabular}
    }
    \caption{{We present a few randomly selected samples of \textbf{audio-to-audio} retrieval.} Here, the frames with {black} borders represent the query, and the next $5$ frames represent the top-5 neighborhoods. The correct retrievals are marked with {{green}}, while the wrong ones are marked with {{red}}. 
    }
    \label{fig:ret_a2a}
\end{figure}

To perform a qualitative analysis of the learned representations in an unsupervised setup, we present the nearest neighborhoods of video-to-video and audio-to-audio retrieval in Figures  \ref{fig:ret_v2v} and \ref{fig:ret_a2a}. In this experiment, we use Kinetics400 \cite{kinetics400} to pretrain CrissCross.
Next, we use the features extracted from randomly selected samples of the validation split to query the training features. We find that in most of the cases CrissCross performs fairly well, we notice very few instances of wrong retrieval, which generally occur when the visual scenes or sound events are very similar. For instance, `playing piano' and `playing organ' for video-to-video retrieval and `playing keyboard' and `playing xylophone' for audio-to-audio retrieval.

\section{Datasets} \label{supsec:datasets}
\subsection{Pretraining Datasets}
We use $3$ datasets of different sizes for pretraining, namely, Kinetics-Sound~\cite{l3-kineticssound}, Kinetics400~\cite{kinetics400}, and AudioSet~\cite{audioset}. Kinetics-Sound is a small-scale action recognition dataset, which has a total of $22$K video clips, distributed over $32$ action classes. Kinetics400 is a medium-scale human action recognition dataset, originally collected from YouTube. It has a total of $240$K training samples and $400$ action classes. Please note that Kinetics-Sound is a subset of Kinetics400, and consists of action classes which are prominently manifested audibly and visually \cite{l3-kineticssound}. Lastly, AudioSet~\cite{audioset} is a large-scale video dataset of audio events consisting of a total of $1.8$M audio-video segments originally obtained from YouTube spread over $632$ audio classes. Please note that none of the provided labels are used in self-supervised pretraining.

\subsection{Downstream Datasets}
Following the standard practices of prior works \cite{avid,ravid,mmv,xdc,selavi,avts}, we evaluate our self-supervised methods on two types of downstream tasks: (\textit{i}) action recognition based on visual representations and (\textit{ii}) sound classification based on audio representations. To perform action recognition, we use two popular benchmarks, i.e., UCF101~\cite{ucf101} and HMDB51~\cite{hmdb}. UCF101 consists of a total of $13$K clips distributed among $101$ action classes, while HMDB contains nearly $7$K video clips distributed over $51$ action categories. To perform sound classification, we use two popular benchmarks ESC50~\cite{esc} and DCASE2014 \cite{dcase}. ESC50 is a collection of $2$K audio events comprised of $50$ classes and DCASE2014 is an audio event dataset of $100$ recordings spread over $10$ categories.


\section{Data Augmentation} \label{supsec:augs}
Here we present the details of the augmentation parameters for both visual and audio modalities.

\subsection{Visual Augmentations}
The parameters for visual augmentations are presented in Table \ref{stab:params_vid_aug}. Some of the parameters are chosen from the literature, while the rest are found through empirical search. We set the parameters of Multi-Scale Crop, Gaussian Blur, and Gray Scale as suggested in \cite{simclr}, and the parameters for Color Jitter are taken from \cite{avid}. We use TorchVision~\cite{Pytorch} for all the implementations of visual augmentations, except Cutout where we use the implementation available here\footnote{\url{https://github.com/uoguelph-mlrg/Cutout}}. Please note that for the Cutout transformation, the mask is created with the mean value of the first frame in the sequence. 

\subsection{Audio Augmentations}
We present the parameters used for audio augmentations in Table \ref{stab:params_aud_aug}. We use the Librosa\cite{librosa} library to generate mel-spectrograms. We use the techniques proposed in \cite{specaug} to perform Time Mask, Frequency Mask, and Time Warp transformations\footnote{\url{https://github.com/s3prl/s3prl}}. 
The parameters for the audio augmentations are set empirically, except for Random Crop which we adopt from \cite{byol_audio}.

\begin{table}[t]
    \fontsize{9pt}{10pt}\selectfont
        \centering
        \begin{tabular}{ll}
        \toprule
        \textbf{Augmentation} & \textbf{Parameters} \\ \midrule\midrule
        {Multi Scale Crop} & \texttt{min area = 0.08} \\ \midrule
        {Horizontal Flip}  & \texttt{p = 0.5} \\ \midrule
        {Color Jitter}     & \texttt{\specialcellleft{brightness = 0.4\\contrast = 0.4\\saturation = 0.4\\hue = 0.2}} \\ \hline
        {Gray Scale}       & \texttt{p = 0.2} \\ \midrule
        {Gaussian Blur}    & \texttt{p = 0.5} \\ \midrule
        {Cutout}           & \texttt{\specialcellleft{max size = 20\\num = 1}} \\
        \bottomrule
        \end{tabular}%
        \caption{Visual augmentation parameters.}
        \label{stab:params_vid_aug}
\end{table}

\begin{table}[t]
        \centering
        \begin{tabular}{ll}
        \toprule
        \textbf{Augmentation} & \textbf{Parameters} \\ \midrule\midrule
        {Volume Jitter}   & \texttt{range = $\pm$0.2} \\ \midrule
        {Time Mask}       & \texttt{\specialcellleft{max size = 20\\num = 2}} \\ \midrule
        {Frequency Mask}  & \texttt{\specialcellleft{max size = 10\\num = 2}} \\ \midrule
        {Timewarp}        & \texttt{wrap window = 20} \\ \midrule
        {Random Crop}     & \texttt{\specialcellleft{range = [0.6,1.5]\\crop scale = [1.0,1.5]}}\\
        \bottomrule
        \end{tabular}%
        \caption{Audio augmentation parameters.}
        \label{stab:params_aud_aug}
\end{table}

\begin{table}[t]
        \centering
        \begin{tabular}{lcccccc}
        \toprule
        & MSC & HF & CJ & GS & GB & C \\ \midrule\midrule
        Pretraining & \cmark & \cmark & \cmark & \cmark & \cmark & \cmark \\
        Full-finetune & \cmark & \cmark & \cmark & \cmark & \xmark & \cmark \\
        Linear evaluation & \cmark & \cmark & \cmark & \cmark & \xmark & \cmark \\ 
        \bottomrule
        \end{tabular}%
        \caption{{Audio augmentation summary.}}
        \label{stab:summ_aud_aug}
        \vspace{10pt}
        \begin{tabular}{lcccc}
        \toprule
        & VJ & Mask & RC & TW \\ \midrule\midrule
        Pretraining & \cmark & \cmark & \cmark & \xmark \\
        Linear evaluation & \cmark & \cmark & \cmark & \cmark \\ 
        \bottomrule
        \end{tabular}%
        \caption{Visual augmentation summary.}
        \label{stab:summ_vid_aug}

\end{table}


\section{Evaluation Protocol} \label{supsec:eval_proto}
To evaluate the representations learned with self-supervised pretraining, we test the proposed framework in different setups, namely  linear evaluation, full finetuning, and retrieval. The details of the evaluation protocols are mentioned below. 

\subsection{Linear Evaluation}
To perform linear evaluations of the learned representations on downstream tasks, we extract fixed features (also called frozen features) using the pretrained backbones. We train a linear classifier using the fixed feature representations. The details are presented below.

\noindent\textbf{Action Recognition.}~\\
To perform linear evaluations on action recognition, we follow standard evaluation protocols laid out in prior works \cite{mmv,brave,gdt,avid}. The details are presented below.

\noindent\textbf{HMDB51 and UCF101.}
We perform linear evaluations in $2$ setups, i.e., $8$-frame and $32$-frame inputs. We evaluate on $8$-frame inputs for the design explorations and $32$-frame inputs for large-scale experiments.

Following the protocols mentioned in \cite{mmv,brave}, we feed $8$-frame inputs to the video backbone, with a spatial resolution of $224^2$. During training, we randomly pick $25$ clips per sample to extract augmented representations, while during testing, we uniformly select $10$ clips per sample and report top-1 accuracy at sample-level prediction by averaging clip-level predictions. The augmentation techniques are mentioned in Section \ref{supsec:augs}. We don't apply the Gaussian Blur while extracting the training features since it deteriorates the performance. Moreover, to perform a deterministic evaluation, we don't apply any augmentations during validation. The visual features are extracted from the final convolution layer and passed to a max-pool layer with a kernel size of $(1, 4, 4)$ \cite{avid}. Finally, we use the learned visual representations to train a linear SVM classifier, we sweep the cost values between $\{$0.00001, 0.00005, 0.0001, 0.0005, 0.001, 0.005, 0.01, 1$\}$ and report the best accuracy. 

When validating on $32$-frame inputs, we could not perform SVM as the feature vector is too large to hold in the memory. Hence, we use a linear fully-connected layer at the end of the video backbone. Note that during training the backbone is kept frozen and only the linear layer is trained. we keep the rest of the setup the same as described earlier, with the exception of training where we randomly select $10$ clips per sample.

\noindent\textbf{Kinetics400.}
As Kinetics400 \cite{kinetics400} is a large-scale dataset, the feature vector is too large to save in memory. Following \cite{avid}, we use a fully connected layer at the end of the frozen backbone and feed $8\times 224^2$ frame inputs. During training, we randomly pick $1$ clip per sample, while during validation, we uniformly select $10$ clips per sample. Note that the rest of the setups remain the same, as described for HMDB51 and UCF101. Finally, we obtain the sample-level prediction by averaging the clip-level predictions and report the top-1 accuracy.

\noindent\textbf{Sound Classification.}~\\
{In case of evaluating audio representations, we follow the evaluation protocol laid out in prior works \cite{avid,xdc,mmv,brave} for respective datasets. The details are mentioned below.
}

\noindent\textbf{ESC50.}
We perform linear evaluations on ESC50 in $2$ setups, we use $2$-second audio input for design exploration and $5$-second audio input for large-scale experiments.
Following \cite{gdt}, we extract $10$ epochs worth of augmented feature vectors from the training clips. During testing, when using $2$-second inputs, we extract $10$ equally spaced audio segments \cite{avid,gdt,xdc}, and when using $5$-second inputs, we extract $1$ segment \cite{mmv,brave} from each sample.  
We perform the augmentations mentioned in Section \ref{supsec:augs} to extract the training features. We notice that unlike self-supervised pretraining, time warping improves the model performance in the linear evaluation. We do not apply any augmentations during validation. We extract the representations from the final convolution layer and pass it through a max-pool layer with a kernel size of $(1, 3)$ and a stride of $(1, 2)$ \cite{gdt}. Similar to action recognition, we perform classification using a one-vs-all linear SVM classifier, we sweep the cost values between $\{$0.00001, 0.00005, 0.0001, 0.0005, 0.001, 0.005, 0.01, 1$\}$ and report the best accuracy.

\noindent\textbf{DCASE.}
To validate on DCASE, we follow the protocol mentioned in \cite{avid}. We extract $60$ clips per sample and train a linear classifier on the extracted representations. Note that the augmentation and feature extraction schemes remain the same as mentioned for ESC50. We report the top-1 sample level accuracies by averaging the clip level predictions. 

\noindent\textbf{Multi-modal Fusion.}
To perform a multi-modal linear evaluation with late fusion, we extract features from Kinetics-Sound. During training, we randomly pick $10$ audio-visual clips per sample, each $2$ seconds long. Next, we extract feature vectors of dimension $2048$ from the last convolution layer by using max-pooling with kernel sizes of $(1, 2, 2)$ and $(1, 4)$ for visual and audio respectively. Following, the feature vectors are concatenated to train a linear SVM classifier. Finally, we report the top-1 sample level accuracy for action classification.

\subsection{Full Finetuning}
Following earlier works \cite{xdc,avid,ravid,selavi}, we use the pretrained visual backbone along with a newly added fully-connected layer for full finetuning on UCF101 \cite{ucf101} and HMDB51 \cite{hmdb}. We adopt two setups for full finetuning, $8$-frame inputs and $32$-frame inputs. In both cases, we use a spatial resolution of $224^2$. Lastly, we replace the final adaptive average-pooling layer with an adaptive max-pooling layer. We find that applying strong augmentations improves the model performance in full-finetuning. Please see the augmentation details in Section \ref{supsec:augs}.
During testing, we extract $10$ equally spaced clips from each sample and do not apply any augmentations. We report the top-1 accuracy at sample-level prediction by averaging the clip-level predictions. We use an SGD optimizer with a multi-step learning rate scheduler to finetune the model. We present the hyperparameters of full-finetuning in Table \ref{stab:params_downstream}.

\subsection{Retrieval}
We follow the protocol laid out in \cite{gdt,cliporder}. We uniformly select $10$ clips per sample from both training and test splits. We fit $2$-second inputs to the backbone to extract representations. We empirically test additional steps such as l2-normalization and applying batch-normalization on the extracted features, and notice that they do not help the performance. Hence, we simply average the features extracted from the test split to query the features of the training split. We compute the cosine distance between the feature vectors of the test clips (query) and the representations of all the training clips (neighbors). We consider a correct prediction if $k$ neighboring clips of a query clip belong to the same class. We calculate accuracies for $k=1,5,20$. We use the NearestNeighbors\footnote{\url{sklearn.neighbors.NearestNeighbors}} API provided in SciKit-Learn in this experiment.


\section{Architecture Details} \label{supsec:arch}
In this study, we use a slightly modified version of R(2+1)D-18 \cite{r2plus1d} as the video backbone as proposed in \cite{avid}, and ResNet-18 \cite{resnet} as the audio backbone. For the sake of completeness, we present the architecture details in Tables \ref{stab:arch_vid_backbone} and \ref{stab:arch_aud_backbone}, respectively. The predictor and projector heads are made of fully-connected layers following \cite{simsiam}, and their architecture details are presented in Table \ref{stab:arch_pred_proj}.

\begin{table}[t]
    \centering

        \resizebox{0.45\textwidth}{!}{%
        \begin{tabular}{rccccccc}
        \toprule
        \bf Layer    & \bf $X_s$ & \bf $X_t$ & \bf $C$ & \bf $K_s$ & \bf $K_t$ & \bf $S_s$ & \bf $S_t$ \\\midrule\midrule
        \bf \texttt{frames}    & 112 & 8 & 3   & -  & - & - & - \\
        \bf \texttt{conv1}    & 56 & 8 & 64  & 7  & 3 & 2 & 1 \\
        \bf \texttt{maxpool} & 28  & 8 & 64  & 3  & 1 & 2 & 1 \\
        \bf \texttt{block2.1.1} & 28  & 8 & 64  & 3  & 3 & 1 & 1 \\
        \bf \texttt{block2.1.2} & 28  & 8 & 64  & 3  & 3 & 1 & 1 \\
        \bf \texttt{block2.2.1} & 28  & 8 & 64  & 3  & 3 & 1 & 1 \\
        \bf \texttt{block2.2.2} & 28  & 8 & 64  & 3  & 3 & 1 & 1 \\
        \bf \texttt{block3.1.1} & 14  & 4 & 128 & 3  & 3 & 2 & 2 \\
        \bf \texttt{block3.1.2} & 14  & 4 & 128 & 3  & 3 & 1 & 1 \\
        \bf \texttt{block3.2.1} & 14  & 4 & 128 & 3  & 3 & 1 & 1 \\
        \bf \texttt{block3.2.2} & 14  & 4 & 128 & 3  & 3 & 1 & 1 \\
        \bf \texttt{block4.1.1} & 7  & 2 & 256 & 3  & 3 & 2 & 2 \\
        \bf \texttt{block4.1.2} & 7  & 2 & 256 & 3  & 3 & 1 & 1 \\
        \bf \texttt{block4.2.1} & 7  & 2 & 256 & 3  & 3 & 1 & 1 \\
        \bf \texttt{block4.2.2} & 7  & 2 & 256 & 3  & 3 & 1 & 1 \\
        \bf \texttt{block5.1.1} & 4   & 1 & 512 & 3  & 3 & 2 & 2 \\
        \bf \texttt{block5.1.2} & 4   & 1 & 512 & 3  & 3 & 1 & 1 \\
        \bf \texttt{block5.2.1} & 4   & 1 & 512 & 3  & 3 & 1 & 1 \\
        \bf \texttt{block5.2.2} & 4   & 1 & 512 & 3  & 3 & 1 & 1 \\
        \bf \texttt{{avg-pool}} & -   & - & 512 & -  & - & - & - \\
        
        \bottomrule
        \end{tabular}
        }
        \caption{{Architecture of the video backbone: R(2+1)D-18.}}
        \label{stab:arch_vid_backbone}
\end{table}

\begin{table}[t]
        \resizebox{0.45\textwidth}{!}{%
        \begin{tabular}{rccccccc}
        \toprule
        \bf Layer    & \bf $X_f$ & \bf $X_t$ & \bf $C$ & \bf $K_s$ & \bf $K_t$ & \bf $S_f$ & \bf $S_t$ \\\midrule\midrule
        \bf \texttt{spectrogram}    & $80$ & $200$ & $1$   & -  & - & - & - \\
        \bf \texttt{conv1}    & $40$ & $100$ & $64$  & $7$  & $7$ & $2$ & $2$ \\
        \bf \texttt{maxpool} & $20$  & $50$ & $64$  & $3$  & $3$ & $2$ & $2$ \\
        \bf \texttt{block2.1.1} & $20$  & $50$ & $64$  & $3$  & $3$ & $2$ & $2$ \\
        \bf \texttt{block2.1.2} & $20$  & $50$ & $64$  & $3$  & $3$ & $2$ & $2$ \\
        \bf \texttt{block2.2.1} & $20$  & $50$ & $64$  & $3$  & $3$ & $2$ & $2$ \\
        \bf \texttt{block2.2.2} & $20$  & $50$ & $64$  & $3$  & $3$ & $2$ & $2$ \\
        \bf \texttt{block3.1.1} & $10$  & $25$ & $128$ & $3$  & $3$ & $2$ & $2$ \\
        \bf \texttt{block3.1.2} & $10$  & $25$ & $128$ & $3$  & $3$ & $2$ & $2$ \\
        \bf \texttt{block3.2.1} & $10$  & $25$ & $128$ & $3$  & $3$ & $2$ & $2$ \\
        \bf \texttt{block3.2.2} & $10$  & $25$ & $128$ & $3$  & $3$ & $2$ & $2$ \\
        \bf \texttt{block4.1.1} & $5$  & $13$ & $256$ & $3$  & $3$ & $2$ & $2$ \\
        \bf \texttt{block4.1.2} & $5$  & $13$ & $256$ & $3$  & $3$ & $2$ & $2$ \\
        \bf \texttt{block4.2.1} & $5$  & $13$ & $256$ & $3$  & $3$ & $2$ & $2$ \\
        \bf \texttt{block4.2.2} & $5$  & $13$ & $256$ & $3$  & $3$ & $2$ & $2$ \\
        \bf \texttt{block5.1.1} & $3$   & $7$ & $512$ & $3$  & $3$ & $2$ & $2$ \\
        \bf \texttt{block5.1.2} & $3$   & $7$ & $512$ & $3$  & $3$ & $2$ & $2$ \\
        \bf \texttt{block5.2.1} & $3$   & $7$ & $512$ & $3$  & $3$ & $2$ & $2$ \\
        \bf \texttt{block5.2.2} & $3$   & $7$ & $512$ & $3$  & $3$ & $2$ & $2$ \\
        \bf \texttt{{avg-pool}} & -   & - & $512$ & -  & - & - & - \\
        \bottomrule
        \end{tabular}
        }
        \caption{{Architecture of the audio backbone: ResNet-18.}}
        \label{stab:arch_aud_backbone}
\end{table}

\begin{table}[t]
    \centering
        \begin{tabular}{rr}
        \toprule
        \bf Layer    & \bf Dimensions \\\midrule\midrule
        \bf \texttt{input}    & 512  \\
        \bf \texttt{fc-bn-relu}    & 2048 \\
        \bf \texttt{fc-bn-relu}    & 2048 \\
        \bf \texttt{fc-bn}    & 2048 \\
        \bottomrule
        \end{tabular}
     \caption{{Architecture of projector heads.}}
     \label{stab:arch_pred_proj}
    \vspace{10pt}

        \begin{tabular}{rr}
        \toprule
        \bf Layer    & \bf Dimensions \\\midrule\midrule
        \bf \texttt{input}    & 2048  \\
        \bf \texttt{fc-bn-relu}    & 512 \\
        \bf \texttt{fc}    & 2048 \\
        \bottomrule
        \end{tabular}
    \caption{{Architecture of predictor heads.}}
    \label{stab:arch_pred}
\end{table}


\section{Hyperparameters and Training Details} \label{supsec:hyperparams}
In this section, we present the details of the hyperparameters, computation requirements, as well as additional training details of self-supervised pretraining and full finetuning. 

\subsection{Pretraining Details}
We present the pretraining hyperparameters of CrissCross in Table \ref{stab:params_pretext}. Most of the parameters remain the same across all $3$ datasets, with the exception of a few hyperparameters such as learning rates and epoch size which are set depending on the size of the datasets. We train on Kinetics-Sound with a batch size of $512$, on a single node with $4$ Nvidia RTX-6000 GPUs. Next, when training on Kinetics400 and AudioSet, we use $2$ nodes and set the batch size to $2048$. Adam \cite{adam} optimizer is used to train our proposed framework. We use LARC\footnote{\url{https://github.com/NVIDIA/apex/blob/master/apex/parallel/LARC.py}}\cite{larc} as a wrapper to the Adam optimizer to clip the gradients while pretraining with a batch size of 2048. In this work, we stick to batch sizes of $512$ and $2048$, because (\textit{i}) as they show stable performance based on the findings of \cite{simsiam}; (\textit{ii}) they fit well with our available GPU setups.
Additionally, we perform mixed-precision training \cite{amp} using PyTorch AMP \cite{Pytorch} to reduce the computation overhead. 

\subsubsection{Ablation Parameters.}~
In the ablation study, we keep the training setup exactly identical across all the variants, with the exception of the learning rates, which we tune to find the best performance for that particular variant. For example, we set the base learning rate for $\loss_{v1v2}$ and $\loss_{a1a2}$models as $0.0001$ and $0.00001$ respectively. Next, the predictor learning rates are set to $0.001$ and $0.0001$ for the $\loss_{v1v2}$ and $\loss_{a1a2}$ variants.

\subsection{Full Finetuning Details}
The full fine-tuning hyperparameters for both benchmarks are presented in Table \ref{stab:params_downstream}. We use a batch size of $32$ for the $32$-frame input and $64$ for the $8$-frame input. We use an SGD optimizer with a multi-step learning rate scheduler to finetune the video backbones. Please note that we perform the full finetuning on a single Nvidia RTX-6000 GPU.

\begin{table*}[t]
\fontsize{9pt}{10pt}\selectfont
\centering

\begin{tabular}{lll}
\toprule
\textbf{Abbreviations} & \textbf{Name} & \textbf{Description} \\ \midrule\midrule
bs & batch size & The size of a mini-batch. \\ \midrule
es & epoch size & The total number of samples per epoch. \\ \midrule
ep & toal epochs & The total number of epochs. \\ \midrule
\specialcellleft{lr\\lr$_{\texttt{ab}}$\\lr$_{\texttt{vb}}$\\lr$_{\texttt{ap}}$\\lr$_{\texttt{vp}}$} & \specialcellleft{learning rate\\audio backbone lr\\video backbone lr\\audio predictor lr\\video predictor lr} & The learning rates to train the networks. \\  \midrule
lrs & learning rate scheduler & The learning rate scheduler to train the network. \\ \midrule
ms & milestones & At every ms epoch the learning rate is decayed. \\ \midrule
$\gamma$ & lr decay rate & The learning rate is decayed by a factor of $\gamma$. \\ \midrule
wd & weight decay & The weight decay used in the SGD optimizer. \\ \midrule
mtm & momentum & The momentum used in the SGD optimizer. \\ \midrule
drp & dropout & The dropout rate. \\ 
\bottomrule
\end{tabular}%
\caption{{Abbreviations and descriptions of the hyperparameters.}}
\label{stab:params_abvt}

\vspace{20pt}
\begin{tabular}{llllllllllll}
\toprule
\textbf{dataset} & \textbf{bs} & \textbf{es} & \textbf{ep} & \textbf{optim} & \textbf{lrs} & \textbf{lr$_\mathbf{vb}$(start/end)} & \textbf{lr$_\mathbf{ab}$(start/end)} & \textbf{lr$_\mathbf{vp}$} & \textbf{lr$_\mathbf{ap}$} & \textbf{wd} & \textbf{betas}   \\ \midrule\midrule
KS & $512$ & $220$K & $100$ & Adam & Cosine & $0.0002/0$ & $0.0002/0$ & $0.002$ & $0.002$ & $0.0001$ & $0.9,0.999$ \\
K400 & $2048$ & $1$M & $100$ & Adam$^*$ & Cosine & $0.0002/0.0001$ & $0.0002/0.0001$ & $0.002$ & $0.002$ & $0.0001$ & $0.9,0.999$ \\
AS & $2048$ & $3.5$M & $100$ & Adam$^*$ & Cosine & $0.0001/0$ & $0.0001/0$ & $0.001$ & $0.001$ & $0.0001$ & $0.9,0.999$ \\
\bottomrule
\end{tabular}%
\caption{{Pretext training parameters. Note the abbreviations used below, KS: Kinetics-Sound, K400: Kinetics400, AS: AudioSet, Adam$^*$: Adam with LARC }} 
\label{stab:params_pretext}

\vspace{20pt}
\begin{tabular}{lllllllllllll}
\toprule
\textbf{dataset} & \textbf{input} & \textbf{es} & \textbf{bs} & \textbf{ep} & \textbf{ms} & \textbf{optim} & \textbf{lrs} & \textbf{lr} & $\mathbf{\gamma}$ & \textbf{wd} & \textbf{mtm} & \textbf{drp}  \\ \midrule\midrule

UCF101 & $8\!\times\!224^2$ & $95$K & $64$ & $20$ & $6/10/14$ & SGD & multi-step & $0.0005$ & $0.3$ & $0.0$ & $0.9$ & $0.0$ \\
UCF101 & $32\!\times\!224^2$ & $95$K & $32$ & $20$ & $8/12/16$ & SGD & multi-step & $0.00007$ & $0.3$ & $0.0$ & $0.9$ & $0.0$ \\
HMDB51 & $8\!\times\!224^2$ & $35$K & $64$ & $20$ & $6/10/14$ & SGD & multi-step & $0.0005$ & $0.1$ & $0.0$ & $0.9$ & $0.0$ \\
HMDB51 & $32\!\times\!224^2$ & $35$K & $32$ & $20$ & $8/12/16$ & SGD & multi-step & $0.0001$ & $0.3$ & $0.0$ & $0.9$ & $0.0$ \\

\bottomrule
\end{tabular}%
\caption{{Full-finetuning hyperparameters for action recognition when pretrained on Kinetics400.}}
\label{stab:params_downstream}
\end{table*}


\section{Limitations.} \label{supsec:limitations}
The notion of asynchronous cross-modal optimization has not been explored beyond audio-visual modalities. For example, our model can be expanded to consider more than $2$ modalities (e.g., audio, visual, and text), which are yet to be studied. Additionally, we notice a considerable performance gap between full-supervision and self-supervision when both methods are pretrained with the same large-scale dataset (Kinetics400 or AudioSet), showing room for further improvement.

\section{Broader Impact.} \label{supsec:borader_impact}
Better self-supervised audio-visual learning can be used for detection of harmful contents on the Internet. Additionally, such methods can be used to develop better multimedia systems. Lastly, the notion that relaxed cross-modal temporal synchronicity is useful, can challenge our existing/standard approaches in learning multi-modal representations and result in new directions of inquiry. The authors don't foresee any major negative impacts.


\end{document}